\documentclass[fleqn,10pt]{wlscirep}
\usepackage[utf8]{inputenc}
\usepackage[T1]{fontenc}
\usepackage{nameref}
\usepackage{amsmath,amsfonts,amssymb,mathtools}
\usepackage{graphicx,float}
\usepackage[ruled,vlined]{algorithm2e}
\usepackage{algorithmic}
\usepackage{xcolor}
\usepackage{cite}
\usepackage{hyperref}
\usepackage{lipsum}
\usepackage{booktabs}
\usepackage{tabularx}
\usepackage{caption}
\usepackage[skip=0.5ex]{subcaption}
\usepackage{multirow}

\title{An Interpretable Hybrid Predictive Model of COVID-19 Cases using Autoregressive Model and LSTM}

\author[1]{Yangyi Zhang}
\author[1,*]{Sui Tang}
\author[2,*]{Guo Yu}
\affil[1]{University of California Santa Barbara, Department of Mathematics, Santa Barbara, CA 93106}
\affil[2]{University of California Santa Barbara, Department of Statistics and Applied Probability, Santa Barbara, CA 93106}

\affil[*]{Corresponding author:suitang@ucsb.edu;guoyu@ucsb.edu}

\begin{abstract} 

The Coronavirus Disease 2019 (COVID-19) has had a profound impact on global health and economy, making it crucial to build accurate and interpretable data-driven predictive models for COVID-19 cases to improve public policy making. 
The extremely large scale of the pandemic and the intrinsically changing transmission characteristics pose a great challenge for effectively predicting COVID-19 cases.
To address this challenge, we propose a novel hybrid model in which the interpretability of the Autoregressive model (AR) and the predictive power of the long short-term memory neural networks (LSTM) join forces. 
The proposed hybrid model is formalized as a neural network with an architecture that connects two composing model blocks, of which the relative contribution is decided data-adaptively in the training procedure.

We demonstrate the favorable performance of the hybrid model over its two single composing models as well as other popular predictive models through comprehensive numerical studies on two data sources under multiple evaluation metrics. Specifically, in county-level data of 8 California counties, our hybrid model achieves 4.173\% MAPE, outperforming the composing AR (5.629\%) and LSTM (4.934\%) alone on average. In country-level datasets, our hybrid model outperforms the widely-used predictive models such as AR, LSTM, Support Vector Machines, Gradient Boosting, and Random Forest, in predicting the COVID-19 cases in Japan, Canada, Brazil, Argentina, Singapore, Italy, and the United Kingdom. In addition to the predictive performance, we illustrate the interpretability of our proposed hybrid model using the estimated AR component, which is a key feature that is not shared by most black-box predictive models for COVID-19 cases.

Our study provides a new and promising direction for building effective and interpretable data-driven models for COVID-19 cases, which could have significant implications for public health policy making and control of the current COVID-19 and potential future pandemics.

\end{abstract}
\begin{document}

\flushbottom
\maketitle

\thispagestyle{empty}

\section*{Introduction}

The coronavirus disease 2019 (COVID-19) pandemic has posed a severe threat to global health and economy while producing some of the richest data we have ever seen in terms of infectious disease tracking. The quantity and quality of data placed epidemic modeling and forecasting at the forefront of worldwide public policy making. Compared to previous infectious diseases, COVID-19 shows special transmission characteristics, yielding significant fluctuations and non-stationarity in the new COVID-19 cases. This poses grand challenges in effective prediction, and, on the other hand, draws attention of the global community to epidemic tracking and prediction.

In the last three years, various models and methods have been developed to predict COVID-19 cases (see survey in \cite{FormerStudy} and references therein).
These models can be roughly grouped into two categories: mechanistic models and data-driven models. The mechanistic models aim at directly characterizing the underlying mechanisms of COVID-19 transmission. Typical examples of mechanistic models are based on differential equations, such as the compartmental models SIR and SEIR\cite{SEIR,bertozzi2020challenges,ndairou2020mathematical,o2022semi}. The data-driven models formulate the prediction of the COVID-19 cases primarily as a regression problem and exploit fully data-adaptive approaches to understand the functional relationship between COVID-19 cases with a set of observable variables. Data-driven models include classical statistical models such as Autoregressive models (AR)\cite{mcdonald2021can,AR,sioofy2021network}, Support Vector Machine (SVM)\cite{GUHATHAKURATA2021351,SVR,tiwari2022pandemic}, and the deep learning models \cite{LSTM,zeroual2020deep,GNN,ALASSAFI2022335,chimmula2020time,hawas2020generated,long2021identification}. In this paper, we will focus on data-driven models.

An Autoregressive model expresses the response variable as a linear function of its previous observations \cite{box2015time}. Its simple structure and strong interpretability are found to be powerful in capturing short-term changing trends in time series. AR models have been applied in various application fields, including infectious decease modeling \cite{AR_Allard, AR_Johansson}. However, they may fail to capture the highly nonlinear patterns and long-term effects in the data-generating dynamics. On the other extreme of the predictive model complexity spectrum, deep learning models, particularly LSTM \cite{hochreiter1997long}, have demonstrate impressive power in capturing complex dependence structures in sequential data. LSTM has been used to achieve the best-known results for many problems on sequential data. However, a well-known limitation of the deep learning models is the short of interpretability due to their black-box nature. This lack of interpretability prevents people from drawing useful conclusions from the model outputs, thus hinders effective policy making\cite{Murdoch1900654116}, especially in crucial fields such as public health. This observation motivates us to consider a hybrid model in which the two seemingly distinct types of models join forces while maintaining both good predictive power and certain interpretability.

In this paper, we propose such a hybrid  model that \textbf{additively combines} the LSTM and the AR model for the task of COVID-19 cases prediction. The proposed hybrid model is formalized as a neural network with an architecture that connects an AR model and a LSTM block, and the relative contribution of these two component models is decided in a totally data-adaptive way in the training procedure.
To demonstrate the predictive power of the proposed hybrid model,
we consider both county-level and country-level data. Specifically, in 8 counties in the state of California, USA, and 7 other different countries (results available in the Supplementary material), our method performs favorably compared with either AR or LSTM model alone, as well as other commonly-used predictive models under various evaluation metrics. All codes are accessible through links on the reference page\cite{git}.

In addition to the predictive accuracy, the importance of predictive models' interpretability has been discussed in plenty of previous works \cite{Murdoch1900654116, Xuhong01756, Vellido04051, Poursabzi07810}. A higher model interpretability facilitates human's ability to understand its predictions, and thus promotes bias detection and other factors that contribute to policy making. Specifically, we demonstrate how the coefficients from the AR part of the trained hybrid model shed light onto understanding the underlying disease transmission mechanism, and thus could help to predict its prevalence trends, and to inform public health policy makers to improve pandemic planning, resource allocation, and implementation of social distancing measures and other interventions. A long-term mission of this paper is to stretch the application of hybrid models beyond COVID-19 forecasting: toward other fast-moving epidemics and cases that require accurate prediction and interpretability.

Although in this paper we focus on confirmed cases prediction, we note that the proposed framework can be easily extended to tackle other COVID-19 or more general epidemiological tasks (e.g., hot spot prediction). Furthermore, the proposed method has its own research significance from a methodological perspective. For example, it raised the open questions on studying its theoretical guarantees, mathematical quantification of prediction, and interpretability.

\subsection*{Related work}

Recently, numerous studies have employed machine learning techniques to investigate various tasks on COVID-19 and achieved impressive results. Examples include using deep learning to detect COVID-19 through CXR images and predicting death status based on food categories to recommend healthy foods during the pandemic\cite{ELARABY2022103441, SHAMS2021104606, Kaur2021}. In light of these advances, our research focuses on predicting confirmed cases of COVID-19.

In this section, we provide a more detailed review of data-driven models that formulate the prediction problem as a regression problem. Regression-based models, including simple AR models and more complex models such as Random Forest, Gradient Boosting, and CNN-LSTM, have been widely used for COVID-19 prediction. For example, Mumtaz et al. \cite{awan2020prediction} used ARIMA to predict the daily confirmed cases in European countries, while Yesilkanat \cite{yecsilkanat2020spatio} used a Random Forest model to predict the number of cases and deaths. Muhammad et al. \cite{muhammad2022cnn} used a CNN-LSTM model to predict the number of confirmed cases and deaths in Nigeria, South Africa, and Botswana. We summarize a list of recent work from year 2020 to 2022 in Table \ref{table:previous_works}.

One advantage of these models is that they do not require a priori knowledge of the disease dynamics and can capture rich relationships in the data. They have been shown to be effective in predicting COVID-19 cases in various regions around the world. However, COVID-19 data displays rich variability, and therefore a single predictive model may not be sufficient and has its own limitations. For example, one major disadvantage of  ARIMA models is that they may not be able to capture non-linear patterns in the data, which can lead to inaccurate predictions. On the other hand, more complex models such as Random Forest and CNN-LSTM may suffer from overfitting, where the model becomes too specialized to the training data and cannot generalize well to new data. These complex models may also lack interpretability, making it difficult to understand the factors driving the predictions and thus provide little to none guidance to actual public health policy making.

Hybrid predictive models that combine different regression models may offer the best of both worlds by capturing both linear and non-linear patterns in the data while maintaining some degrees of interpretability. The idea is to decompose a model into different components that are designed to capture specific characteristics of the data. It has proven to be an effective way of improving empirical predictions in various applications, including those in COVID-19 prediction \cite{ijerph13080757, WeirongYan201063.264, zhao2022prediction,ala2021modeling,zheng2020predicting}.

\subsubsection*{Comparison to previous works on hybrid modeling} 

The idea of using an additive combination of AR models (or more generally, ARIMA models) with LSTMs has recently appeared in the literature for time series forecasting, with applications in gas and oil well production and sunspot monitoring \cite{fan2021well, fathi2019time}. However, there is a significant difference between our approach and previous methods: our approach trains the two components in the model jointly, while previous hybrid modeling techniques take a sequential approach to training.
Specifically, Zhang \cite{ZHANG2003159} proposed a hybrid model of ARIMA and artificial neural networks, aiming to capture more patterns in the data and improve forecasting performance. The preprocessed data is used to fit an ARIMA model first, before the residual term is used as input to train a neural network model. Fan et al. \cite{fan2021well} followed a similar procedure, using an ARIMA model and an LSTM model. The logistics of these methods is to use an ARIMA model to capture the linear pattern of the data first and rely on the neural networks capture the non-linear pattern in the residuals. The main goal of these previous works is to explore whether a hybrid model produces better performance than the single models.

In our study, we design a general network architecture that includes both an AR part and an LSTM part additively and trains the entire architecture jointly by minimizing the empirical risk. By doing so, we do not arbitrarily give preference to any of the two additive components. Instead, the relative weights of the interpretable AR part and the predictive LSTM part are determined fully by the data.

\begin{table}[htbp]
\centering
\resizebox{\textwidth}{!}{\begin{tabular}{@{}lllll@{}}
\toprule
  Study &
  Dataset &
  Models &
  Results \\ \midrule\\
  Atik, 2022\cite{Atik5823} &
  Turkey & \begin{tabular}[c]{@{}l@{}}Support Vector Machine (SVM),\\ Linear Regression,\\ Bagged Tree, Fine Tree
  \end{tabular} & \begin{tabular}[c]{@{}l@{}}SVM has the highest R value (99\%) and the lowest\\ RMSE and MAE values.\end{tabular}\\\\
  Galasso et al., 2022\cite{GALASSO2022111779} &
  USA counties & Random Forest & \begin{tabular}[c]{@{}l@{}}MAE below 300 weekly cases/100,000,  using easily\\ accessible training features \end{tabular}\\\\
  Ali et al., 2022\cite{Abdulwahab2022} &
  Saudi Arabia & \begin{tabular}[c]{@{}l@{}}
   Gradient Tree Boosting,\\ Random Forest, XGBoost,\\ Voting Regressor
  \end{tabular} & \begin{tabular}[c]{@{}l@{}} The XGBoost and Voting Regressor model outperform \\the other models\end{tabular}\\\\
  Chumachenko et al., 2022 \cite{computation10060086} &
  \begin{tabular}[c]{@{}l@{}}Germany, Japan,\\ South Korea, and Ukraine\end{tabular} & \begin{tabular}[c]{@{}l@{}} Random Forest, K-Nearest Neighbors,\\
  Gradient Boosting
  \end{tabular} & \begin{tabular}[c]{@{}l@{}} The Gradient Boosting model has the best performance\\ by related error and MAE\end{tabular}\\\\
  Fang et al., 2022 \cite{Fange056685} &
  USA & ARIMA, XGBoost & \begin{tabular}[c]{@{}l@{}} The XGBoost model outperforms the ARIMA\\
  model on COVID-19 case prediction in the USA\end{tabular}\\\\
  Muhammad et al., 2021 
  \cite{Muhammads42979} &
  Mexico &
  \begin{tabular}[c]{@{}l@{}}Logistic Regression, \\ Decision Tree, \\ SVM, Naive Bayes, and \\ Artificial Neutral Network\end{tabular} &
  \begin{tabular}[c]{@{}l@{}}Decision tree model has the highest accuracy (94.99\%).\\ SVM model has the highest sensitivity (93.34\%).\\ Naive Bayes model has the highest specificity (94.30\%).\end{tabular}\\\\
  Luo et al., 2021 \cite{LUO2021104462} & America & double-layer LSTM, XGBoost & 
  \begin{tabular}[c]{@{}l@{}} MAPE of LSTM and XGBoost reaches 2.32\% and 7.21\%
  \end{tabular}\\\\
  Vadyala et al., 2021 \cite{VADYALA2021100085} &
  USA, State of Louisiana & LSTM, SEIR & 
  \begin{tabular}[c]{@{}l@{}} RMSE of LSTM reaches 601.20 while the SEIR\\ model has an RMSE of 3615.83 (on 3 Parishes\\ of Louisiana)
  \end{tabular}\\\\
  Alassafi et al., 2021 \cite{ALASSAFI2022335} &
  \begin{tabular}[c]{@{}l@{}}Malaysia, Morocco and Saudi Arabia\end{tabular} & LSTM, RNN & 
  \begin{tabular}[c]{@{}l@{}} LSTM models reaches a 98.58\% precision accuracy\\ while the RNN models have 93.45\%
  \end{tabular}\\\\
  Tomar and Gupta, 2020 \cite{TOMAR2020138762} &
  India &
  LSTM &
  \begin{tabular}[c]{@{}l@{}}LSTM's prediction closely matches the official data.\\ By the prediction of the LSTM, social isolation and\\ lockdown can reduce the spread of virus effectively.\end{tabular}\\\\
  Bhandari et al., 2020 \cite{BhandariPPR184651} & India & ARIMA &
  \begin{tabular}[c]{@{}l@{}}ARIMA(3,3,1) model has the best performance for\\ cumulative cases and death(predicted RMSE is\\ 668.7 and base RMSE is 5431).\end{tabular}\\\\
  \bottomrule\\
\end{tabular}}
\caption{A non-exhaustive list of previous works on data-driven models for COVID-19 cases prediction in the past three years. Most commonly used evaluation metrics are RMSE, MAE, and MAPE.}
\label{table:previous_works}
\end{table}

\noindent In summary, our contributions can be summarized as:
\begin{itemize}
\item 
Development of a novel approach to hybrid modeling for COVID-19 cases prediction: we have designed a general network architecture that combines AR and LSTM models additively and trains the entire architecture jointly, allowing the relative weights of the interpretable AR part and the predictive LSTM part to be fully determined by the data. This approach is a departure from traditional sequential modeling approach and has the potential to contribute to the literature of sequential data prediction.

\item Extensive numerical studies on data sets from two sources that displays a rich variety of variability: we have shown that the proposed hybrid model demonstrates better forecasting performance than single models. This finding is important as it shows that the hybrid model is an effective way to combine the strengths of different modeling techniques and can be used as a framework for future research.

\item Exploration of interpretability: we have also explored the interpretability of the hybrid model, which is an important contribution as it allows for a better understanding of the model and can lead to improved decision-making based on the model's output. This contribution enhances the practical applicability of our proposed hybrid model.
\end{itemize}

\section*{Methods}

\label{sec:methods}
\noindent In this section, we first overview the two building blocks of our additive hybrid model, namely the AR and the LSTM model, and their relative advantages. Then we present our  hybrid model which combines these two building blocks additively, and we intuitively elaborate why it is better than the two individual components.

\subsection*{Autoregressive (AR) models} 

\begin{figure}[!htb]
    \centering
    \includegraphics[width=0.6\linewidth]{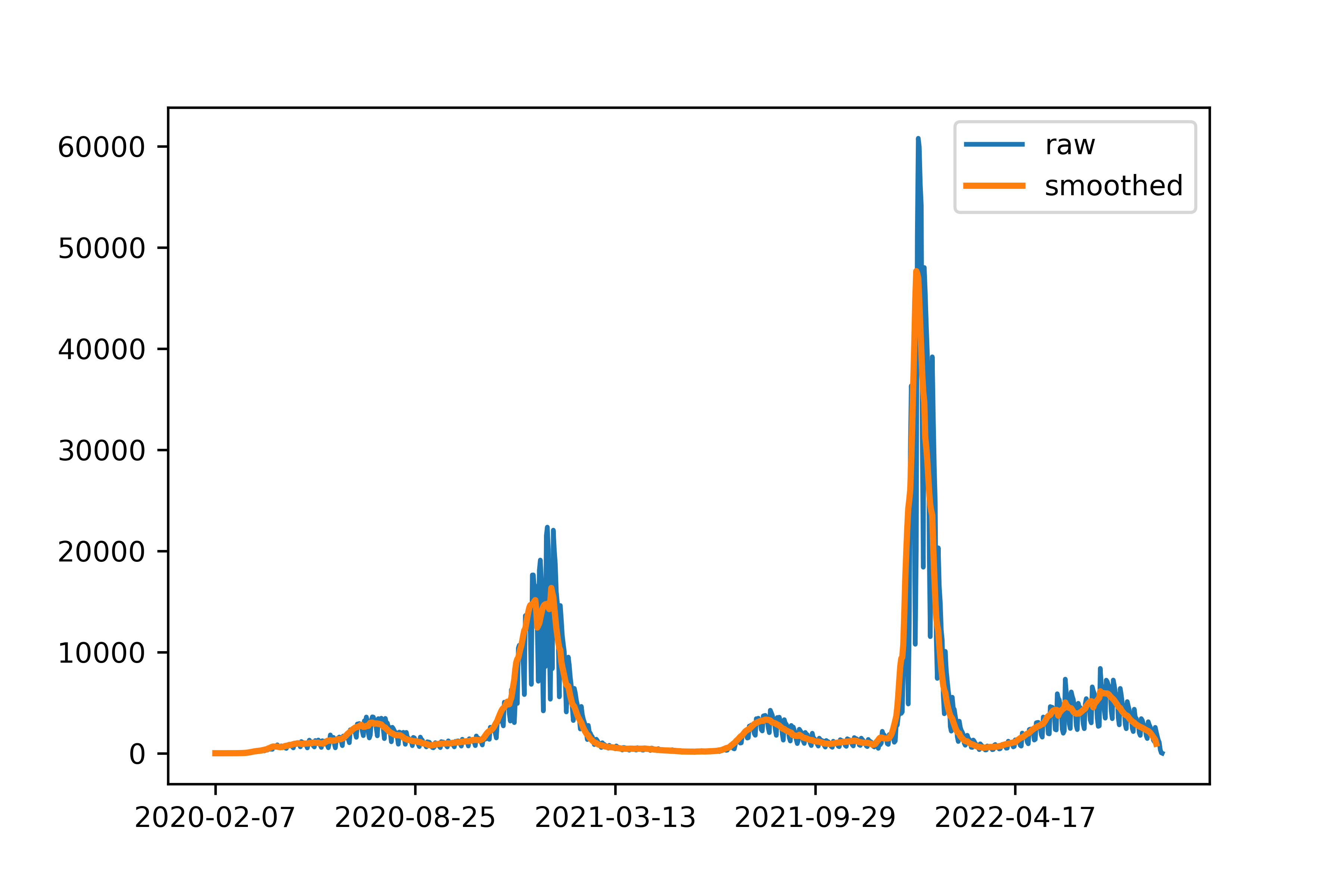}
    \caption{An example of visualizing daily observations, where blue line represents the data before smoothing, orange line represents data after smoothing. The data is collected from the Los Angeles county.}\label{fig:nonlinearity}
\end{figure}

In time series, we often observe associations between past and present values. For example, by knowing the price of a stock in the past few days, we can often make a rough prediction about its value tomorrow. AR is a simple model that utilized this empirical observation and can yield very accurate prediction in certain applications. It represents the time series values using linear combination of the past values. The number of past values used is called the lag number and often denoted by $p$. Let $\epsilon_t$ denote the Gaussian noise at time $t$ with mean 0 and variance $\sigma^2$. The structure equation of AR(p) model can be represented as 

\begin{align}
Y_t =a_0+a_1 Y_{t-1}+a_2Y_{t-2}+\cdots+a_p Y_{t-p}+\epsilon_t
\end{align} 
 where $a_0$ is the intercept, and $a_1,\cdots,a_p$ represent the coefficients. 
\noindent AR model is often effective on stationary data. To ensure stationarity, a common trick is to apply the differencing operation on the time series. A time series value at time $t$ that has been differenced once, $Y^{(1)}$, is defined as follows: 
\begin{align}\label{differencing}
    Y^{(1)}_t= Y_t-Y_{t-1},
\end{align} and higher order differencing operation can be defined recursively.
However, an AR model is not sufficient to capture the non-linear dependence structure, which is found to be an important feature of the COVID-19 data, indicated by Figure \ref{fig:nonlinearity}.
A purely AR based model is thus often  insufficient for the task of COVID-19 cases prediction. 

\subsection*{Long short term memory networks (LSTM)}
RNN (Recurrent Neural Network) \cite{Lipton00019} is known to suffer from the long term dependency problem: as the network grows larger through time, the gradient decays quickly during back propagation, making it impossible to train RNN models with long unfolding in time. To solve this problem, Hochreiter and Schmidhuber (1997) introduced a special type of RNN called LSTM with a proper gradient-based learning algorithm \cite{hochreiter1997long}.

\begin{figure}[H]
    \centering
    \includegraphics[scale=0.18]{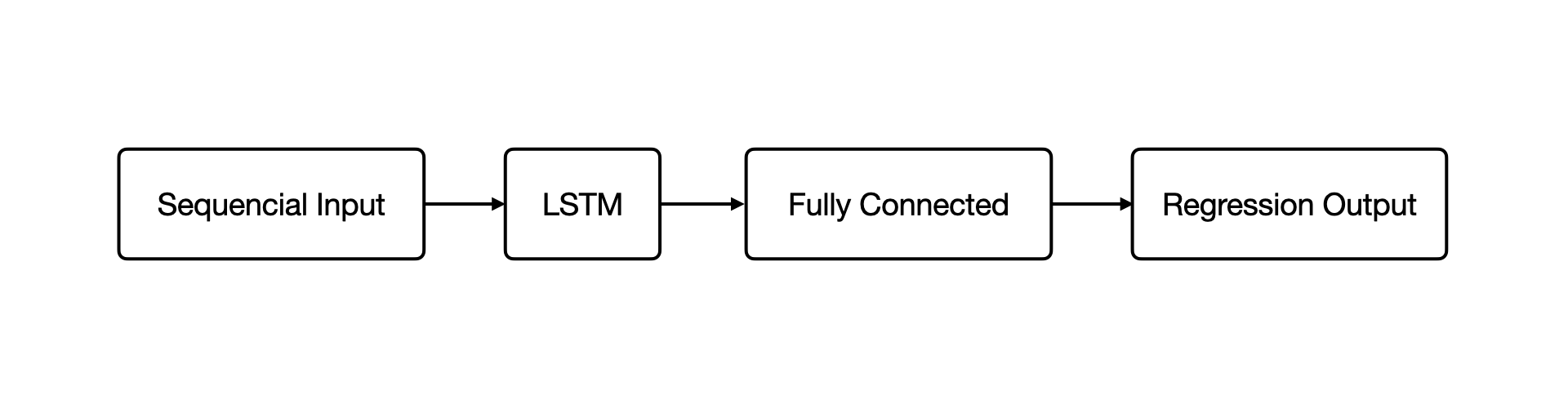}
    \vspace{-0.5em}
    \renewcommand{\figurename}{Supplementary figure}
 \caption{This diagram illustrates the architecture of a simple LSTM network for regression. The network starts with a sequence input layer followed by an LSTM layer. The network ends with a fully connected layer and a regression output layer.}
    \label{fig:LSTM_structure.png}
\end{figure}

We employ a LSTM regression model, which is represented as \begin{equation}
    Y_t = G_{\theta}(Y_{t-1},...,Y_{t-p}),
\end{equation} where we use $Y_{t-1},...,Y_{t-p}$ as the sequential input data; $G$ represents the neural network architecture shown in Figure 
\ref{fig:LSTM_structure.png} and $\theta$ represents the weight parameters in neural networks.

\begin{figure}[H]
    \centering
    \includegraphics[scale=0.2]{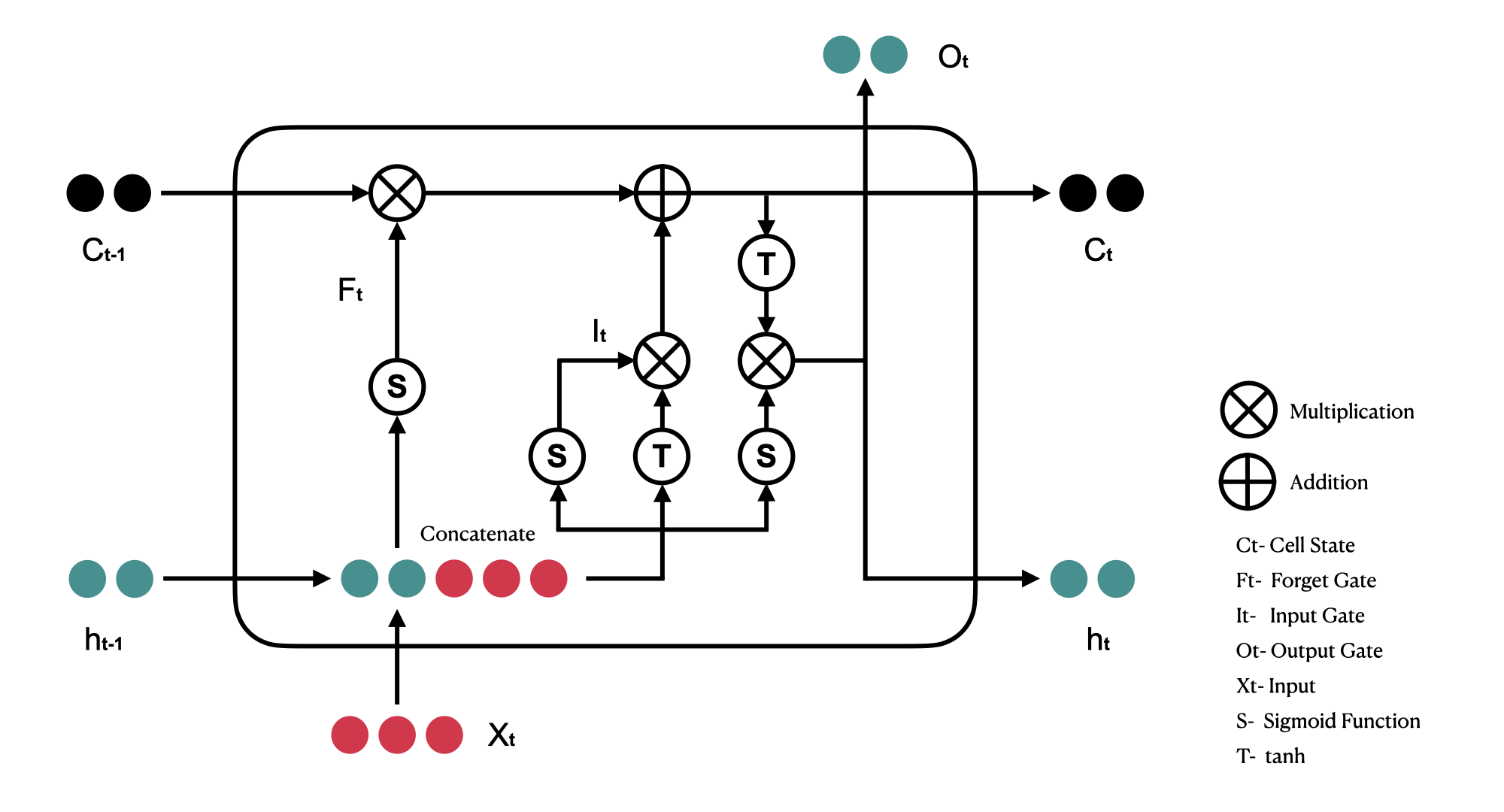}
    \renewcommand{\figurename}{Supplementary figure}
    \caption{Structure of a LSTM cell that describes the cell state $c_t$ (green circles, dim =2) and hidden state $h_t$ (red circles, dim =2) and various gates \cite{LSTMCELL}.}
    \label{fig:LSTMCELL}
\end{figure}

The core concepts of a LSTM cell are the cell states and the associated gates, as illustrated in Figure \ref{fig:LSTMCELL}. The cell state $C_{t-1}$ at time step $t-1$ acts as a transport highway that transfers relative information all the way down the sequence chain, which intuitively characterizes the ``memory'' of the network. 
The cell states, in principle, carry relevant information throughout the processing of the sequence. 
So even information from the earlier time steps can make its way to later time steps, reducing the effects of short-term memory.
The Forget Gate decides what information should be kept. 
The Input Gate decides what information is to be added from the current step and update the cell state $C_t$ at time step $t$.
The Output Gate determines what the next hidden state $h_t$ should be. The four gates comprise a fully connected feed forward neural network.

To achieve optimal prediction results using LSTM model, it is crucial to have a careful hyperparameter tuning, including the choice of units (dimension of the hidden state), the number of cells (i.e. the number of time steps), and layers. This is usually a difficult task in practice. For example, few LSTM cells are unlikely to capture the structure of the sequence, while too many LSTM cells might lead to overfitting.
However, just like other neural networks, a well-known limitation of LSTM is its lack of interpretability \cite{Murdoch1900654116}. 

\subsection*{The hybrid model}
As discussed above, both AR and LSTM have their relative strength and limitations in their prospective domains. 
We propose to combine the two models additively into one single hybrid model, which is expressed as 
\begin{equation}\label{equation}
    Y_t =  \alpha\mathrm{AR}(p) + (1-\alpha) G_{\theta}(Y_{t-1},...,Y_{t-p}),
\end{equation}
where $p$ is the lag number and $\alpha$ weights the contribution of two components: by tuning the value of $\alpha$, one can strike a balance between the prediction given by AR and LSTM parts, and thus a prediction of linear and nonlinear signals.



We illustrate the structure of the hybrid model in Figure \ref{fig:hybrid}. The hybrid model is characterized as one neural network architecture where the two composing models are added through the last layer. The AR component captures the linear relationship in time series and the LSTM component would describe the nonlinear patterns. In section Training of the Supplementary material, we show how to train the weights in each of the two components in a fully data-adaptive manner by minimizing the empirical risk. We will compare the contribution of the hybrid model's AR component and LSTM component in section Results.

\begin{figure}
    \centering
    \includegraphics[scale=0.22]{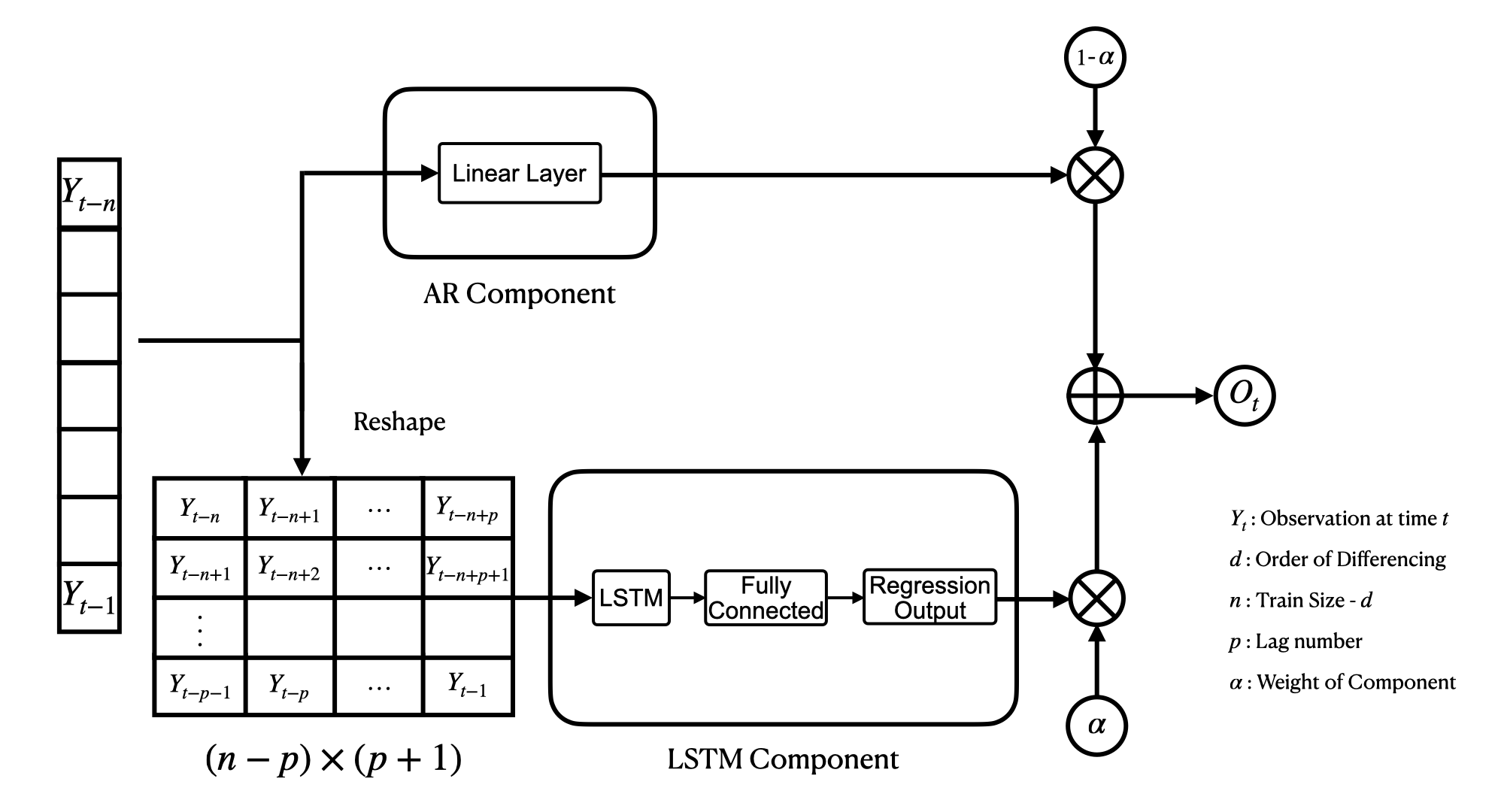}
    \caption{Visualization of the hybrid architecture.}
    \label{fig:hybrid}
\end{figure}

\section*{Results}\label{sec:results}

The results include four sections: Model evaluations, Prediction,  Interpretability, and Comparative study on the WHO datasets. In Model evaluations, we introduce the metrics we use to evaluate the models and on which we compare the models' performances.
In section Prediction, we exhibit the visualizations of several interesting trials and compare the numerical predictions and evaluations of the three models. In the Interpretability part, we compare the AR component of the hybrid model with the AR model. This is to examine how we may interpret the hybrid model. We leave other training details in the Supplementary material. In section Comparative study on the WHO datasets, we further examine the performance of the proposed hybrid model by applying it to data of 7 different countries around the world and comparing its performance with that of its component models and 3 additional models.

\subsection*{Data description and statistical analysis}
We utilize two primary data sources. The first data source is a dataset specific to California counties, which is available in the CHHS Open Data repository under the title \href{https://data.ca.gov/dataset/covid-19-time-series-metrics-by-county-and-state}{COVID-19 Time-Series Metrics by County and State}. This dataset includes information on populations, positive and total tests, number of deaths, and positive cases. We conducted a preliminary statistical analysis to examine correlations between these variables and the number of daily cases. The results of this analysis can be found in Supplementary figure \ref{fig:correlation.png}
in the Supplementary material, and we anticipate that they will provide valuable insights for future research.

The second data source, used for comparative analysis, can be found in the WHO repository at the \href{https://covid19.who.int/data}{WHO Coronavirus (COVID-19) Dashboard}. This resource presents official daily counts of COVID-19 cases, deaths, and vaccine utilization, as reported by countries, territories, and areas. In this study, we use 7 countries: Japan, Canada, Brazil, Argentina, Singapore, Italy, and the United Kingdom.

All datasets generated and analysed during the current study are also available in the \href{https://github.com/Yangyi-Zhang/Covid-Forecasting/tree/main/data}{author's Github repository} \cite{git}.

\subsection*{Model evaluations}
\label{sec:metrics}
We use a quantitative measure to evaluate and compare the performance of models: the Mean Absolute Percentage Error (MAPE), defined as:
\begin{align}
    \mathrm{MAPE} &= \frac{100}{n}\sum_{t=1}^{n}\frac{ |\widehat{Y}_{t} - Y_{\mathrm{true}, t}|}{|Y_{\mathrm{true}, t}|}
\end{align} 
A model with small values of MAPE is preferred.

We examine the performance of the three models (hybrid, AR, and LSTM) on different time periods within the available range. This is essential in our research, since the performance of a model is not constant on different trends; by intuition, a model performs better on smooth curves than it does on steep curves. By repeating our evaluation process on different time periods thus different trends, we wish to understand what trends do the model give the best performance. Such understanding will help us decide to what degrees we may trust the performance of the models. 
We evaluate the models repeatedly to reduce the influence brought by the instability of model training.
Specifically, we leave 7 days between the first date of any two consecutive training data points. Although a larger number of repetitions seems desirable, increasing the repetition number is at the cost of making neighboring training points closer to each other. However, the difference in performance between two neighboring training points, that are too close to each other, would be attributed more to the instability of model training than to the difference in trend. Such results give us little information about the model performance over trend. In the end, we let the step number be the same as our lag number. By doing so, we suppose the concept of a week is important in forecasting.

\subsubsection*{Additional evaluation metrics} 
In the Supplementary material, we additionally evaluate and compare above models using Root Mean Square Error (RMSE) and Mean Absolute Error (MAE). The evaluation is done on the same dataset across different comparing methods.

\begin{figure}[!tbp]
    \centering
    \begin{subfigure}{1\textwidth}
        \centering
        \begin{minipage}{.375\linewidth} \centering \small \hspace{2mm} San Diego Raw Data from 2020-12-03 to 2021-02-28\end{minipage}
        \begin{minipage}{.575\linewidth} \centering \small \hspace{2mm} Prediction versus Ground Truth\end{minipage}
        \includegraphics[width=.375\linewidth, height=0.25\linewidth
        ]{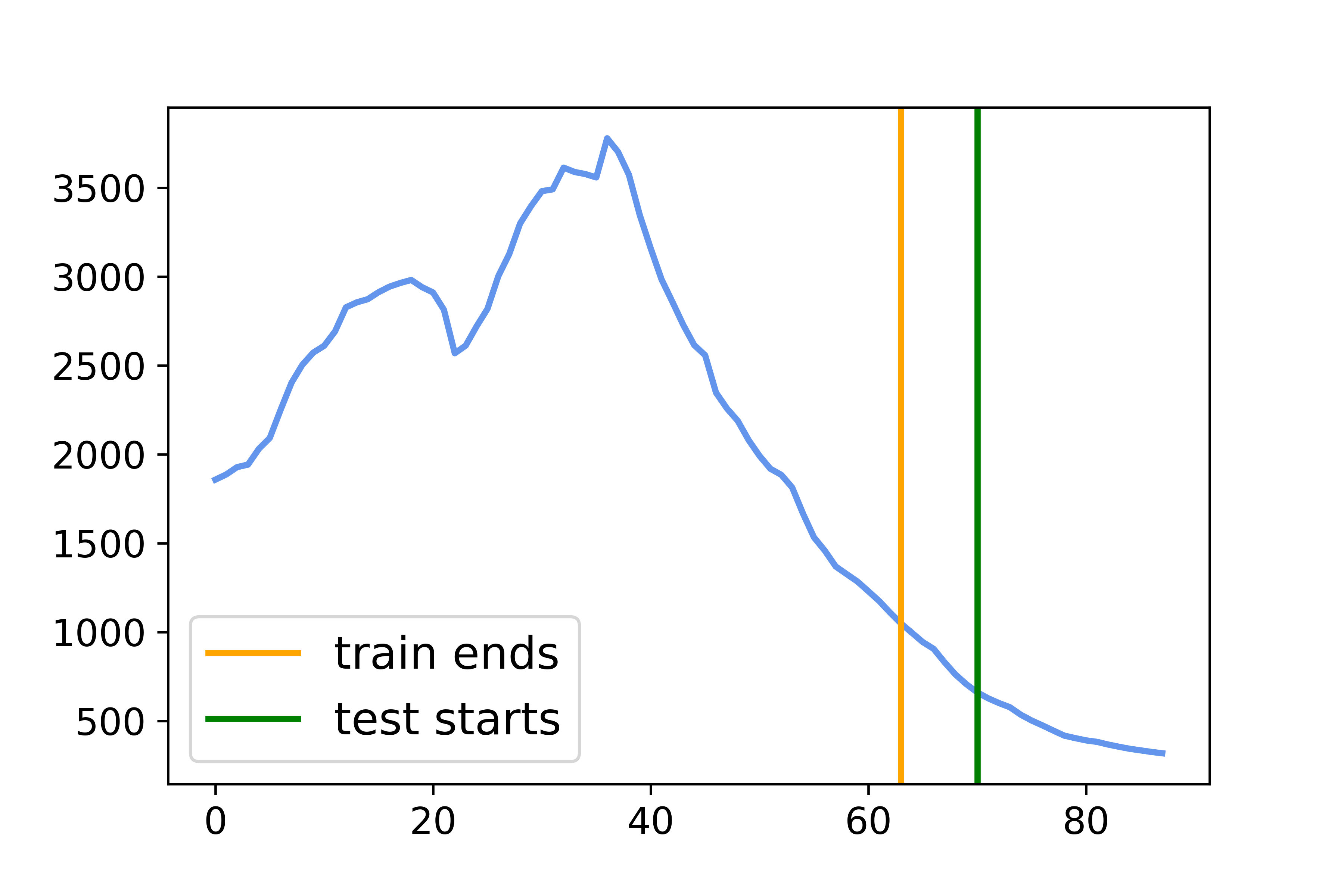}
        \includegraphics[width=.575\linewidth, height=0.25\linewidth
        ]{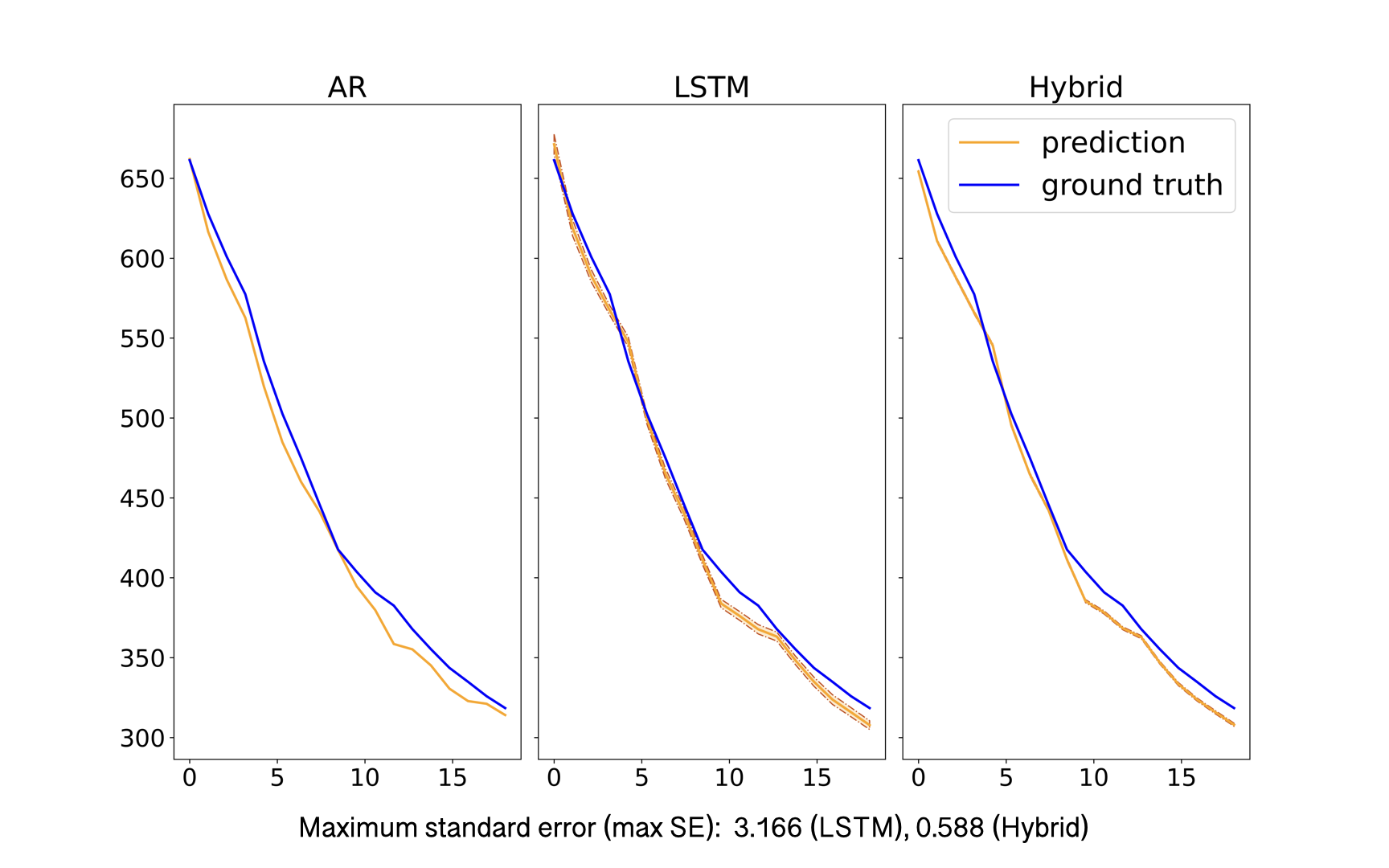}
        \subcaption{Case 1. Curved Training Data and Down Trend Testing Data}
        \label{fig:visual_case1}
    \end{subfigure}
    \begin{subfigure}{1\textwidth}\ContinuedFloat
        \centering
        \begin{minipage}{.375\linewidth} \centering \small \hspace{2mm} San Francisco Raw Data from 2020-02-17 to 2020-05-14\end{minipage}
        \begin{minipage}{.575\linewidth} \centering \small \hspace{2mm} Prediction versus Ground Truth: AR, LSTM, and hybrid (in order)\end{minipage}
        \includegraphics[width=.375\linewidth, height=0.25\linewidth
        ]{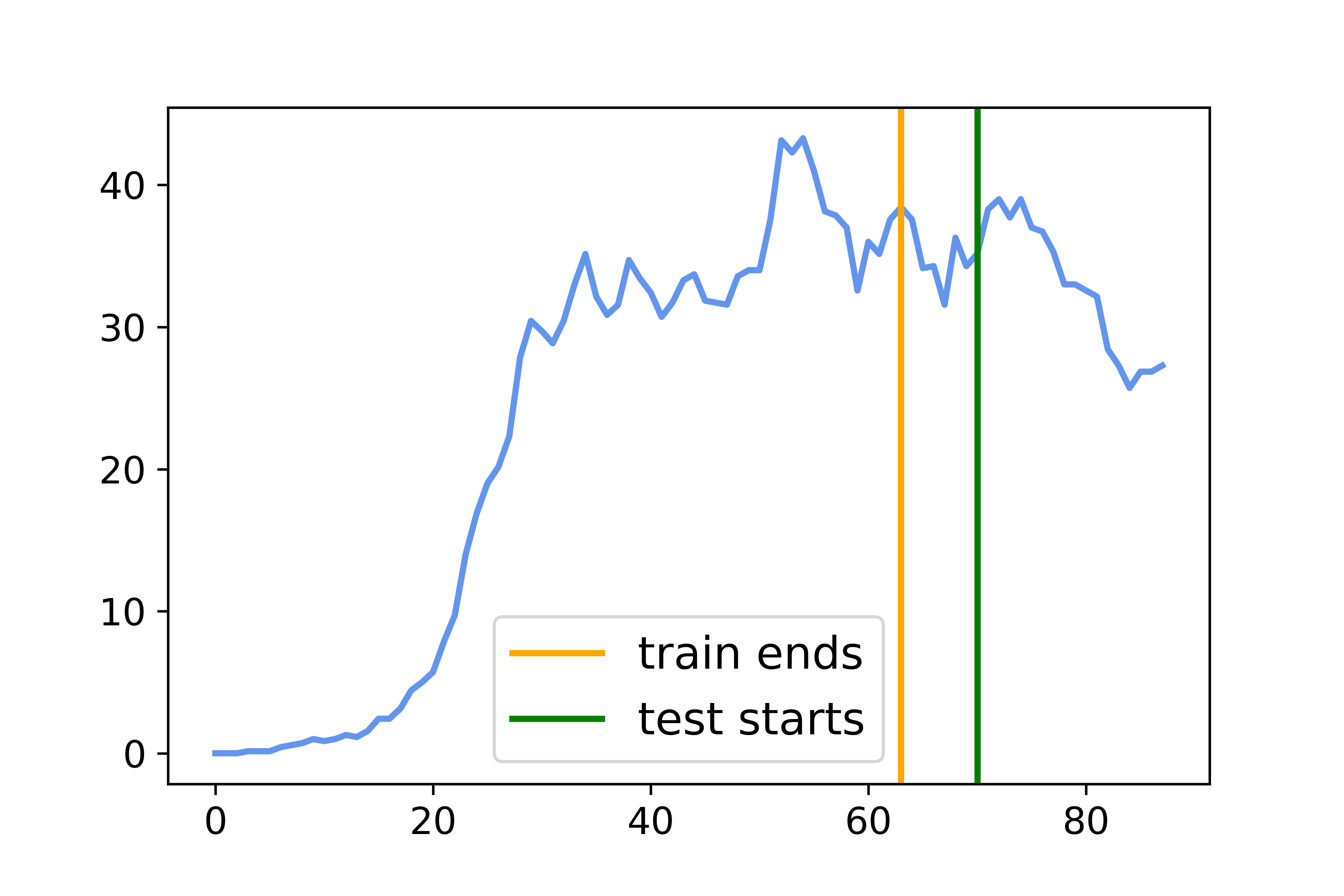}
        \includegraphics[width=.575\linewidth, height=0.25\linewidth
        ]{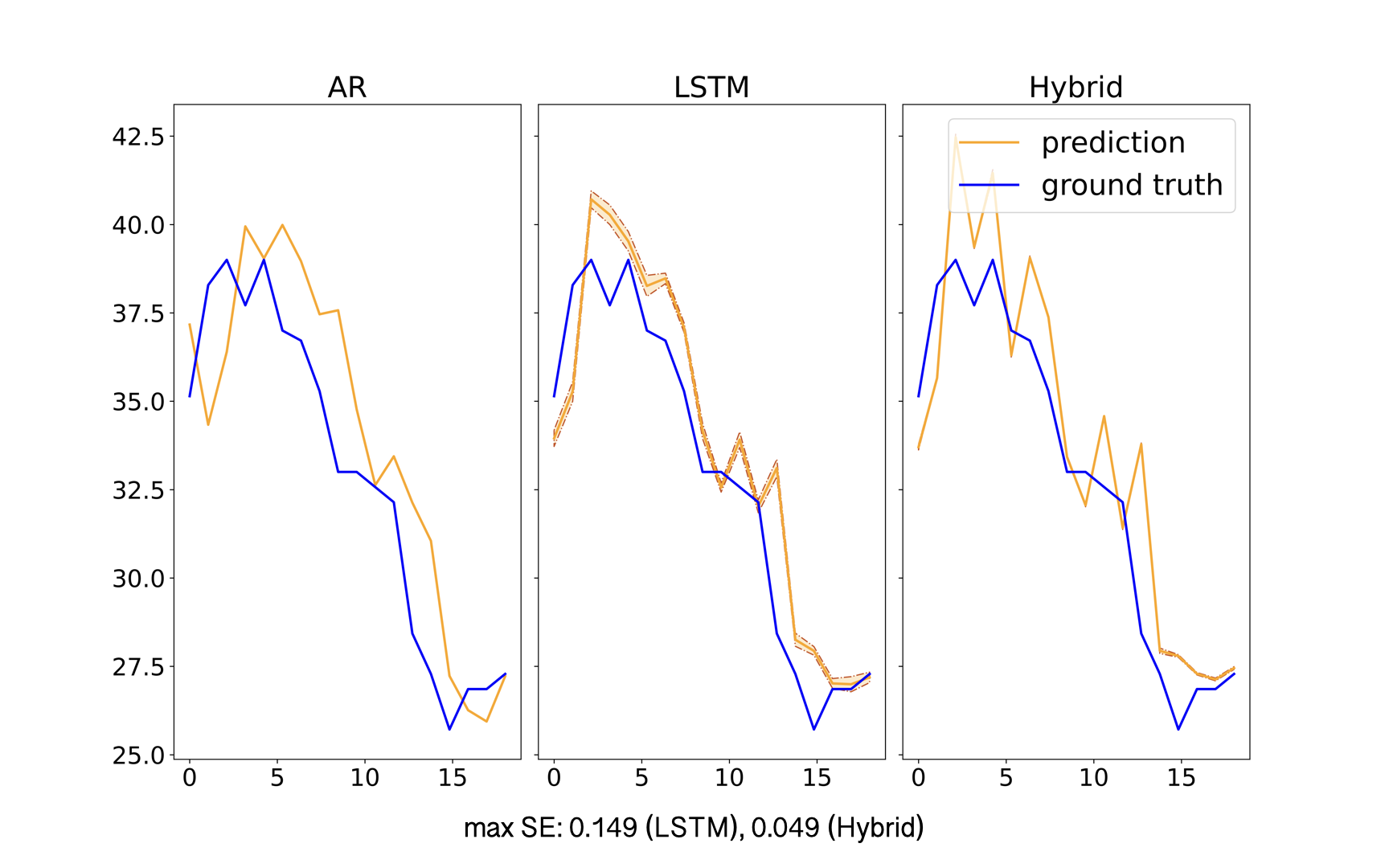}
        \subcaption{Case 2. Up Trend Training and Down Trend Testing}
        \label{fig:visual_case2}
    \end{subfigure}
    \medskip
    \begin{subfigure}{1\textwidth}
        \centering
        \begin{minipage}{.375\linewidth} \centering \small \hspace{2mm} Los Angeles Raw Data from 2020-09-24 to 2020-12-20\end{minipage}
        \begin{minipage}{.575\linewidth} \centering \small \hspace{2mm} Prediction versus Ground Truth\end{minipage}
        \includegraphics[width=.375\linewidth, height=0.25\linewidth
        ]{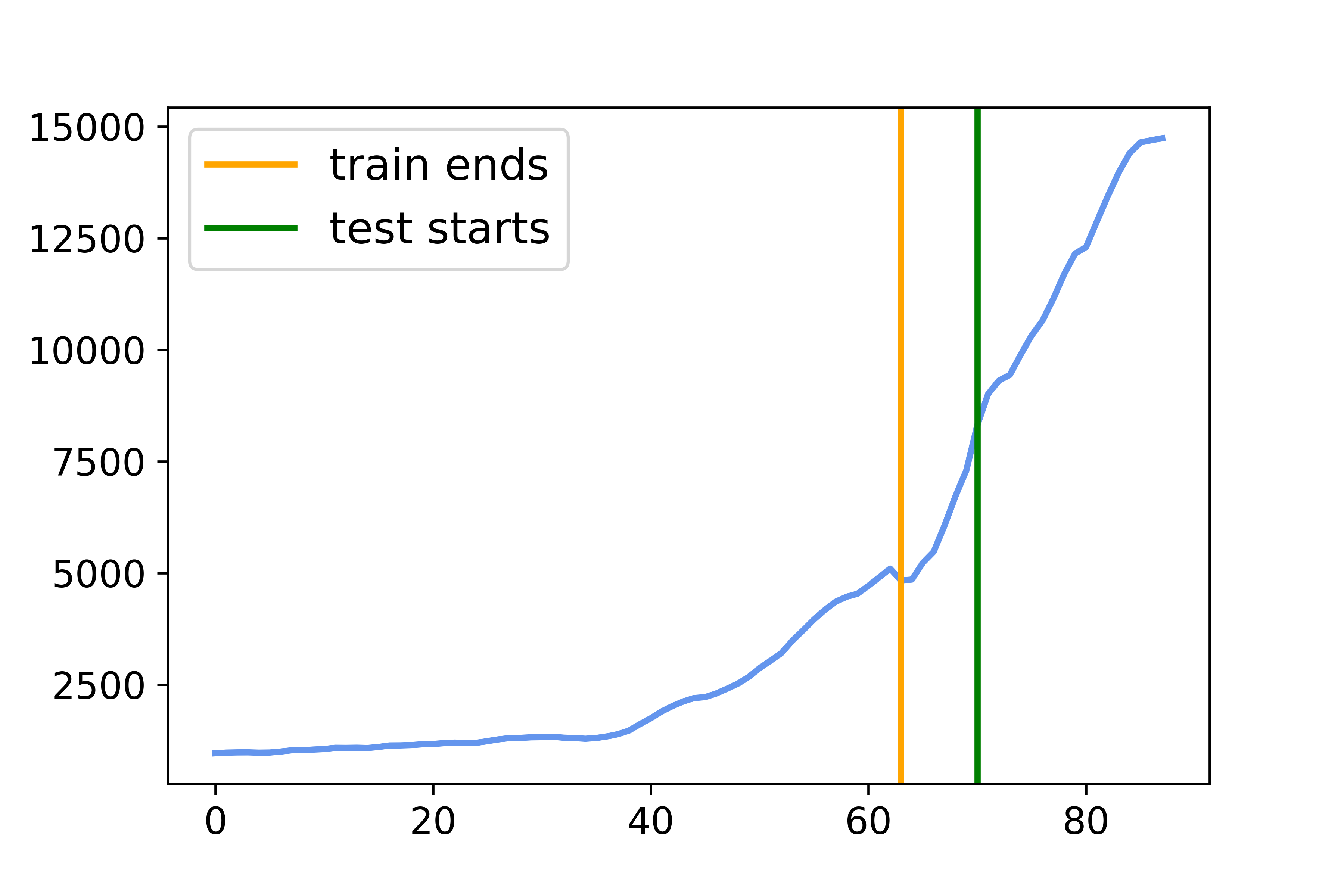}
        \includegraphics[width=.575\linewidth, height=0.25\linewidth
        ]{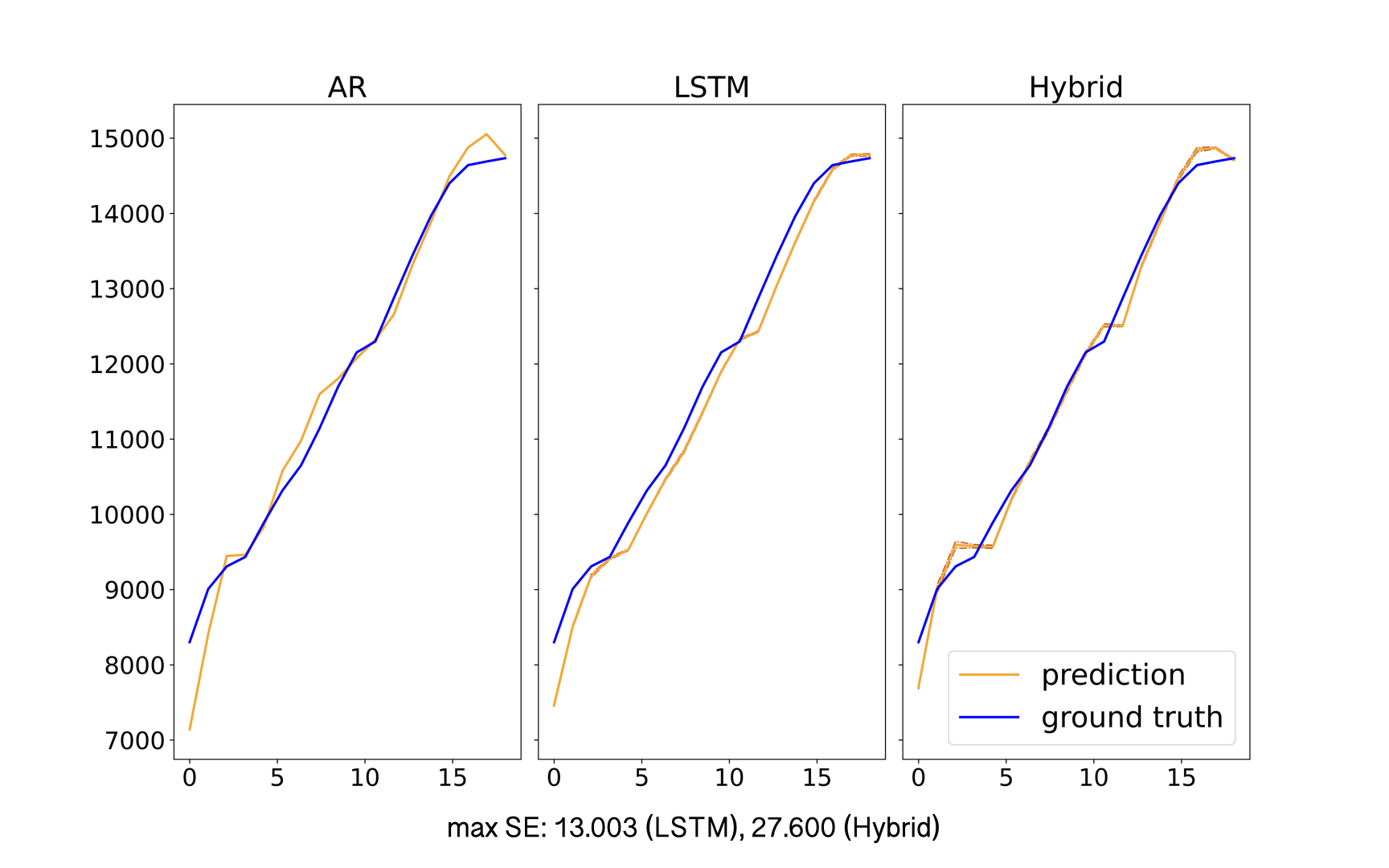}
        \subcaption{Case 3. Up Trend Training and Up Trend Testing}
        \label{fig:visual_case3}
    \end{subfigure}
    \begin{subfigure}{1\textwidth}
        \centering
        \begin{minipage}{.375\linewidth} \centering \small \hspace{2mm} San Francisco Raw Data from 2022-06-10 to 2022-09-05\end{minipage}
        \begin{minipage}{.575\linewidth} \centering \small \hspace{2mm} Prediction versus Ground Truth\end{minipage}
        \includegraphics[width=.375\linewidth, height=0.25\linewidth
        ]{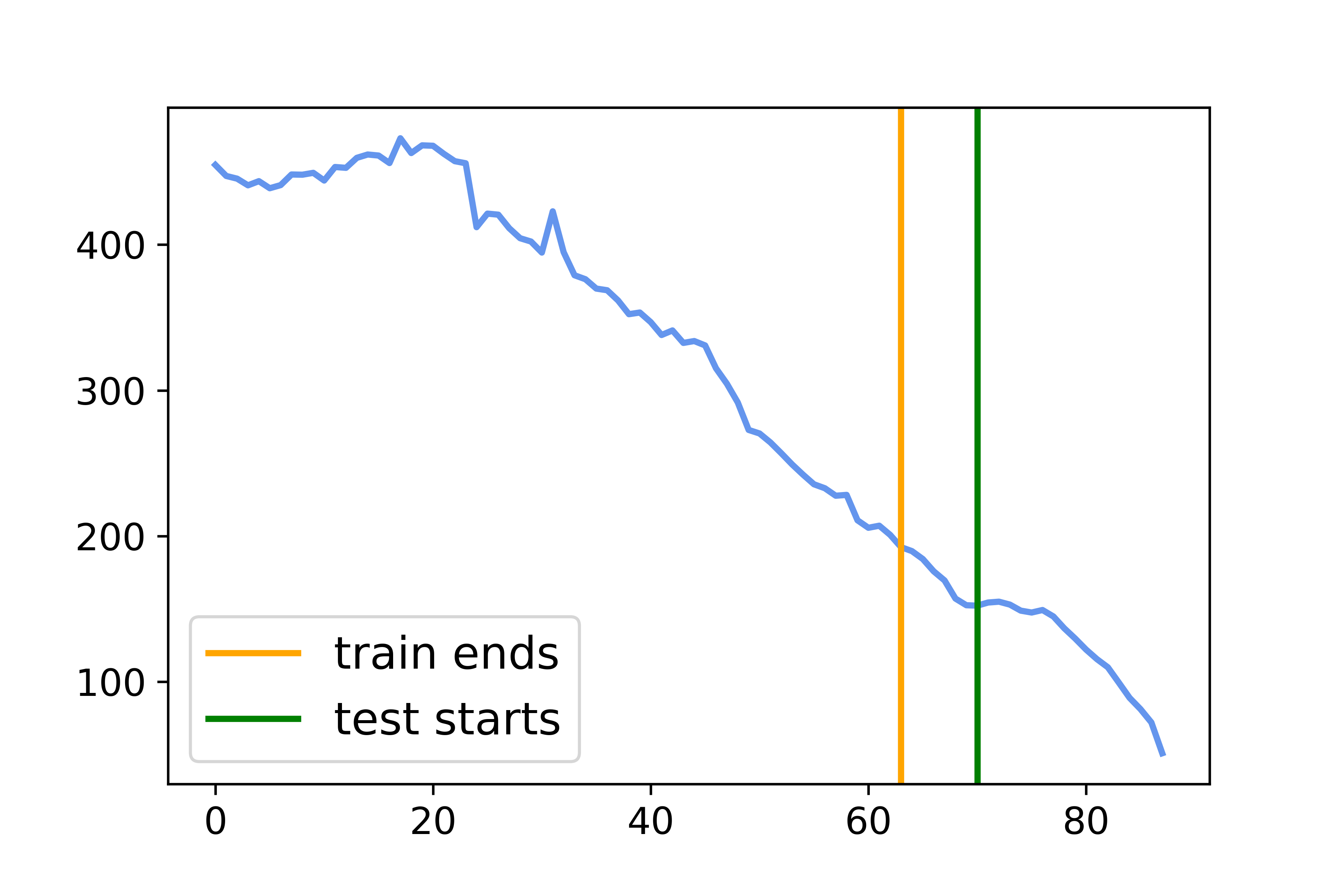}
        \includegraphics[width=.575\linewidth, height=0.25\linewidth
        ]{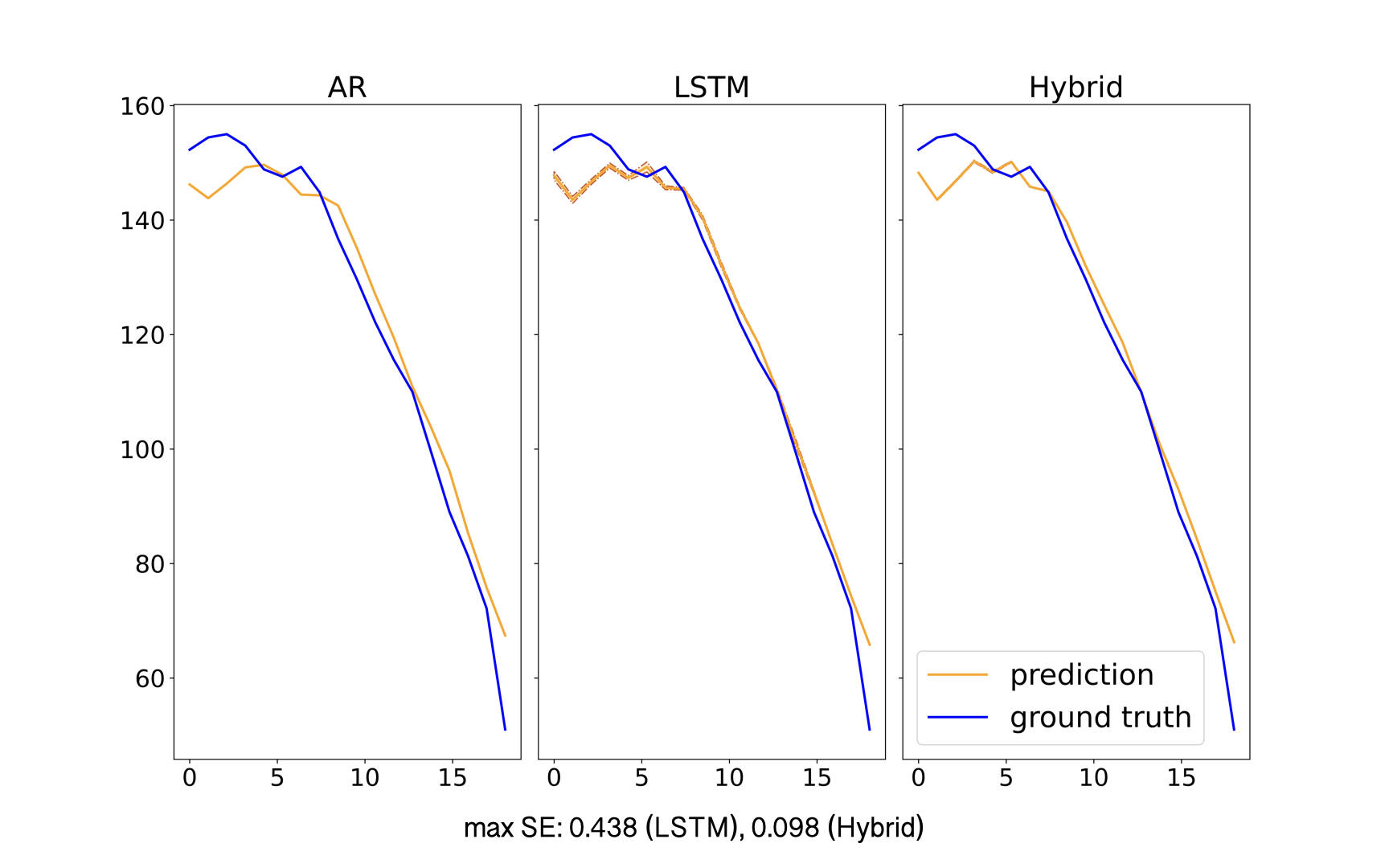}
        \subcaption{Case 4. Down Trend Training and Down Trend Testing}
        \label{fig:visual_case4}
    \end{subfigure}
    \caption{The left panels show the training and testing data. The right panels show the ground truth versus forecasts of the AR, LSTM, and hybrid model, respectively. We display the average prediction (solid line) with 2 times standard error (shaded region). The standard error across 100 runs are reported for LSTM and hybrid. The hybrid model is more stable than the LSTM.} \label{fig:visualization_1}
\end{figure}

\subsection*{Prediction}

In this section, we present the visual and numerical results for all three models. We perform a comprehensive comparison of the performance for the three models in multiple counties, showing the advantage of the hybrid model. All predictions are transformed back to the original scale.

\subsubsection*{Visualization}
\label{sec:Visualizations}

We compare the three models' performance on COVID-19 case prediction in California 8 counties. For each county, we test the models' performance on several different situations: for example, when the training data has an up trend and the testing data has a down trend. From all trials we practiced, we choose the following trials, presented in Figure \ref{fig:visualization_1} and Figure \ref{fig:visualization_2}, as representatives of different combinations of training and testing data, since they reflect the general model performances well. 

Figure \ref{fig:visual_case1} shows models being trained on curved data and being tested on down trend data, as shown on the left and right panel, respectively. Figure \ref{fig:visual_case2} shows models being trained on up trend data and being tested on down trend data. Figure \ref{fig:visual_case3} shows models being trained on up trend data and being tested on up trend data. Figure \ref{fig:visual_case4} shows models being trained on down trend data and being tested on down trend data. Figure \ref{fig:visual_case51} and Figure \ref{fig:visual_case52} show models being trained on down trend data and being tested on up trend data, while Figure \ref{fig:visual_case51} has gentle upward testing data and Figure \ref{fig:visual_case52} has sharp upward testing data. Figure \ref{fig:visual_case6} show models being trained and tested on jagged data.

To ensure the results above are representative, we run each selected trial 100 times, visualize the mean and standard error of these trials, and present averaged MAPE. While AR outperforms LSTM on some cases, the hybrid model outperforms both in most cases, except that in Figure \ref{fig:visual_case2} and in Figure \ref{fig:visual_case6}. The MAPE, averaged on the 100 trials, shows that LSTM (4.469\%) outperforms hybrid (4.993\%) slightly in Figure \ref{fig:visual_case2}. However, as shown in the right panel of Figure \ref{fig:visual_case2}, the hybrid model captures the general trend of ground truth better than LSTM does. Similarly, in Figure \ref{fig:visual_case6}, AR (3.675\%) outperforms hybrid (3.718\%) slightly. Yet, as shown in the right panel of Figure \ref{fig:visual_case6}, the hybrid model captures the general trend of ground truth better than AR does.

Beside, interestingly enough, the hybrid  model always seems to capture the ground truth's trend. Actually, the shape of hybrid 's forecasts resembles either that of the AR model or that of the LSTM model, or it resembles a combination of both. When AR model captures the trend better than the LSTM does, the hybrid  model resembles the AR model in forecast shape: for example, in Figure \ref{fig:visual_case2}, San Francisco 2020-02-17 to 2020-05-14, and in Figure \ref{fig:visual_case51}, Santa Barbara 2022-01-17 to 2022-04-14. When LSTM model captures the trend better than the AR does, the hybrid  model resembles the LSTM model in forecast shape: for example, in Figure \ref{fig:visual_case4}, San Francisco 2022-06-10 to 2022-09-05, and in Figure \ref{fig:visual_case52}, Riverside 2022-02-16 to 2022-12-20. On jagged testing data, where AR performs better on some part and LSTM better on the other, the hybrid model presents advantages of both models: for example, in Figure \ref{fig:visual_case6}, the hybrid model resembles AR on the two ends, where AR performs better, and it resembles LSTM in shape between day 5 to day 15, where LSTM seems to capture the trend better.

\subsubsection*{General performance}
We evaluated the model performances numerically, in the 8 California counties across multiple trials. The results are given in Table \ref{table:performance}. We observe that the hybrid model outperforms the AR model and the LSTM models almost uniformly: it generally yields the smallest average MAPE. To be specific, the general MAPE of each model (AR, LSTM, LSTM with 2 layers, and hybrid), averaged on the results for all 8 counties, is 5.629\%, 4.934\%, 6.804\%, and 4.173\% in order. In general, the hybrid model has the best general performance, and it outperforms the AR model by approximately 1.5\%. The LSTM model suffers from overfitting when a second LSTM layer is added.
As seen in the Supplementary material, the proposed hybrid model also yields the lowest RMSE and MAE values.

\begin{table}
\centering
\begin{tabular}{|l|l|l|l|l|l|}
\hline
& County & AR & LSTM & LSTM (Double) & hybrid \\ \hline
\multirow{8}{*}{General performance} & Los Angeles & 5.153 (1.071) & 5.120 (1.358) & 7.838 (3.414) & \textbf{3.640} (0.773) \\ \cline{2-6} 
&San Diego & 4.446 (0.652) & 3.828 (0.669) & 5.539 (1.549) & \textbf{3.462} (0.883) \\ \cline{2-6} 
& San Francisco & 4.754 (0.421) & 4.433 (0.600) & 5.350 (1.029) & \textbf{3.553} (0.317) \\ \cline{2-6} 
& Santa Barbara & 7.706 (1.714) & 6.157 (1.087) & 8.482 (2.677) & \textbf{5.970} (1.243) \\ \cline{2-6} 
& Fresno & 6.549 (1.668) & 5.092 (0.933) & 6.811 (2.189) & \textbf{4.188} (0.707) \\ \cline{2-6} 
& Sacramento & 5.240 (0.787)  & 4.707 (0.758) & 5.784 (1.182)  & \textbf{3.897} (0.733) \\ \cline{2-6} 
& Ventura & 6.525 (0.706) & 5.607 (0.913) & 7.157 (1.626) & \textbf{4.687} (0.733) \\ \cline{2-6} 
& Riverside & 5.660 (0.925) & 4.528 (0.651) & 7.467 (2.428) & \textbf{3.985} (0.750) \\ \hline
\multirow{8}{*}{Latest performance} & Los Angeles & 4.221 & 3.607 (0.105) & 3.528 (0.161) & \textbf{3.356} (0.030) \\ \cline{2-6} 
& San Diego & 3.749 & 3.513 (0.055) & 3.542 (0.119) & \textbf{3.343} (0.019) \\ \cline{2-6} 
& San Francisco & 5.251 & 4.115 (0.076) & \textbf{4.068} (0.068) & 4.104 (0.028) \\ \cline{2-6} 
& Santa Barbara & 4.731 & 4.183 (0.034) & \textbf{4.111} (0.061) & 4.160 (0.011) \\ \cline{2-6} 
& Fresno & 3.475 & 3.107 (0.067) & 3.525 (0.219) & \textbf{2.942} (0.009) \\ \cline{2-6} 
& Sacramento & 4.685 & 3.789 (0.048) & \textbf{3.703} (0.066) & 3.880 (0.025) \\ \cline{2-6} 
& Ventura & 4.143 & 3.567 (0.029) & 3.489 (0.053) & \textbf{3.410} (0.021) \\ \cline{2-6} 
& Riverside & 4.752 & 3.415 (0.100)  & 3.283 (0.145) & \textbf{3.197} (0.044) \\ \hline
\end{tabular}
\caption{MAPE (by percentage) for each model on each county. General performance is averaged on all trials. The inconsistent performances of neural networks have been compensated by the small step value, which is 7. The Latest performance is on the latest trial, from 2022-06-10 to 2022-09-05. The results for LSTM, LSTM double and hybrid are each averaged on 100 runs, to compensate the inconsistent performances of neural networks. The value in parenthesis is the standard error. AR has 0 or small standard error for the same trial. The hybrid model is usually the best in performance and has the lowest standard error. Best performances are given in bold.} \label{table:performance}
\end{table}

\begin{figure}
    \centering
    \begin{subfigure}{1\textwidth}\ContinuedFloat
        \centering
        \begin{minipage}{.375\linewidth} \centering \small \hspace{2mm} Santa Barbara Raw Data from 2022-01-17 to 2022-04-14\end{minipage}
        \begin{minipage}{.575\linewidth} \centering \small \hspace{2mm} Prediction versus Ground Truth\end{minipage}
        \includegraphics[width=.375\linewidth, height=0.25\linewidth
        ]{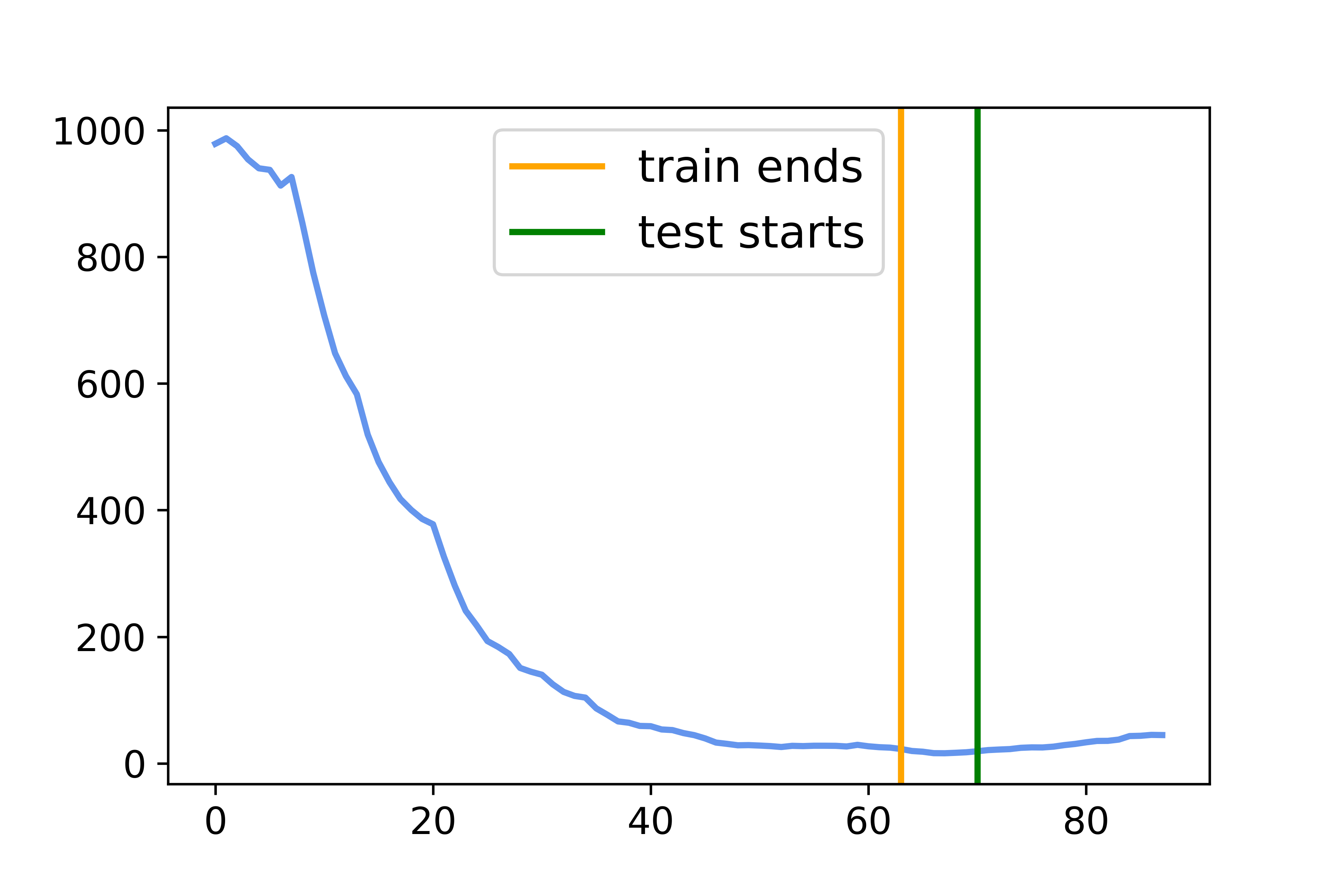}
        \includegraphics[width=.575\linewidth, height=0.25\linewidth
        ]{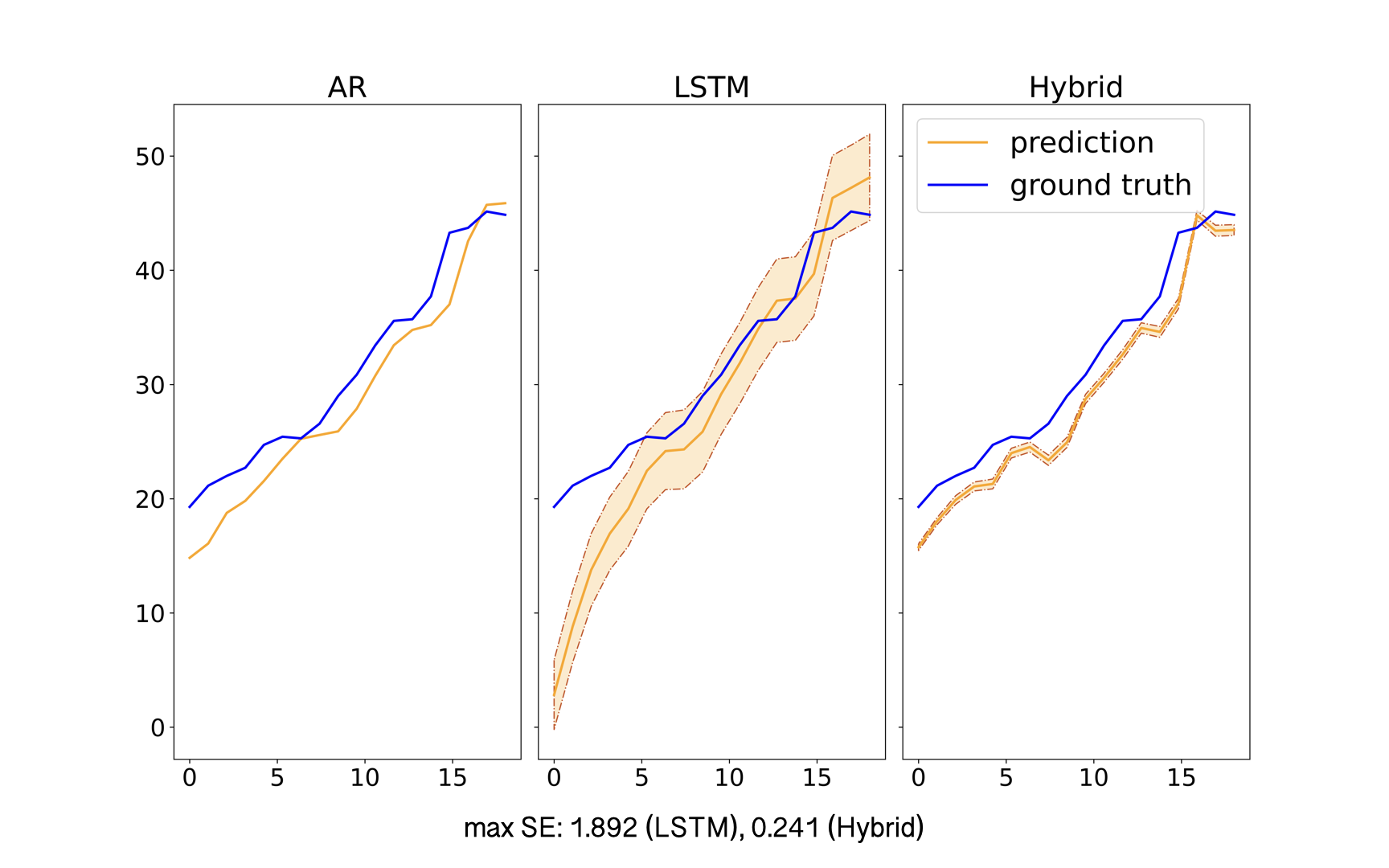}
        \subcaption{Case 5. Down Trend Training and Up Trend Testing}
        \label{fig:visual_case51}
    \end{subfigure}
    \medskip
    \begin{subfigure}{1\textwidth}
        \centering
        \begin{minipage}{.375\linewidth} \centering \small \hspace{2mm} Riverside Raw Data from 2022-02-16 to 2022-05-14\end{minipage}
        \begin{minipage}{.575\linewidth} \centering \small \hspace{2mm} Prediction versus Ground Truth\end{minipage}
        \includegraphics[width=.375\linewidth, height=0.25\linewidth
        ]{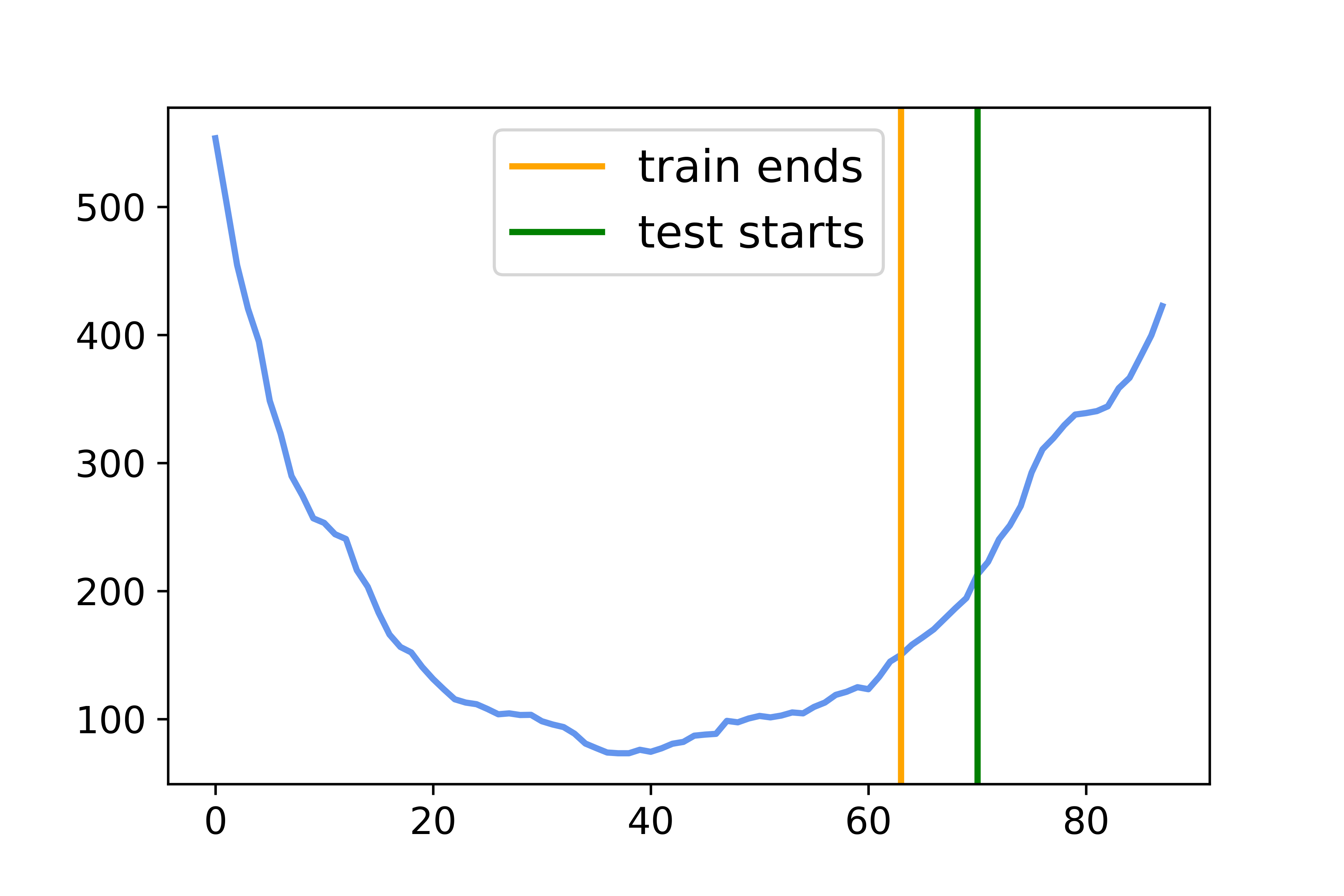}
        \includegraphics[width=.575\linewidth, height=0.25\linewidth
        ]{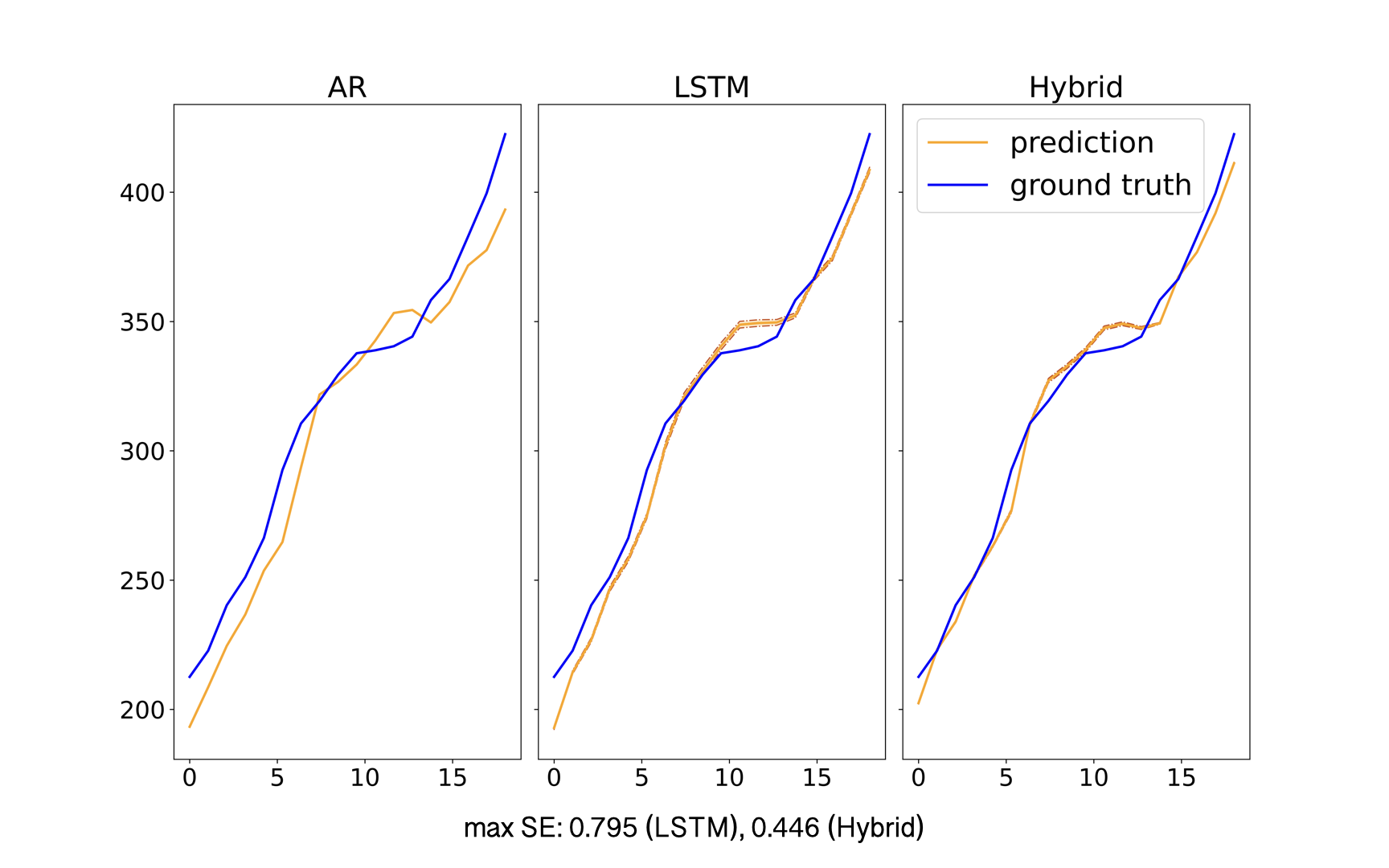}
        \subcaption{Case 5. Down Trend Training and Up Trend Testing}
        \label{fig:visual_case52}
    \end{subfigure}
    \begin{subfigure}{1\textwidth}
        \centering
        \begin{minipage}{.375\linewidth} \centering \small \hspace{2mm} Fresno Raw Data from 2021-02-11 to 2021-05-09\end{minipage}
        \begin{minipage}{.575\linewidth} \centering \small \hspace{2mm} Prediction versus Ground Truth\end{minipage}
        \includegraphics[width=.375\linewidth, height=0.25\linewidth
        ]{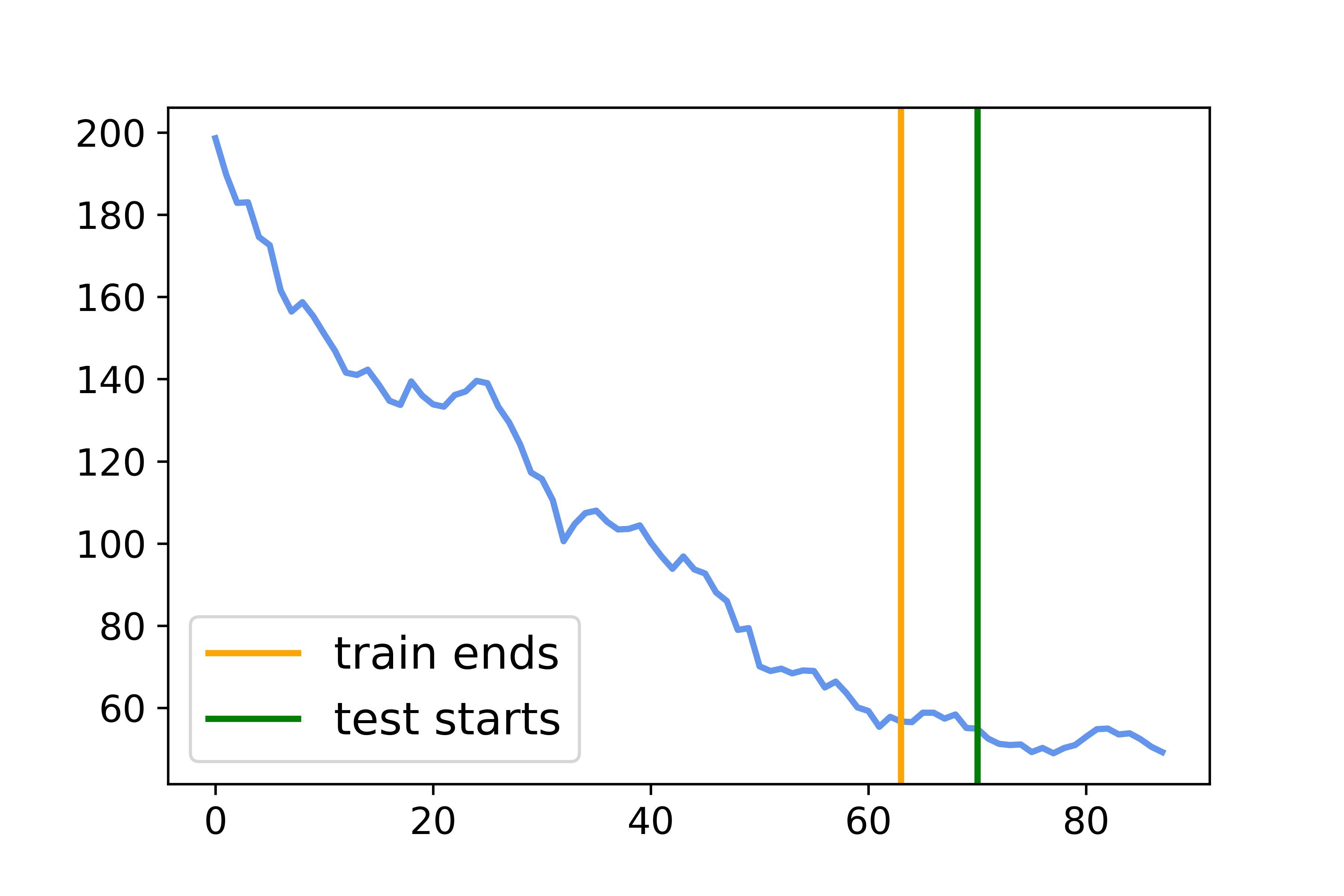}
        \includegraphics[width=.575\linewidth, height=0.25\linewidth
        ]{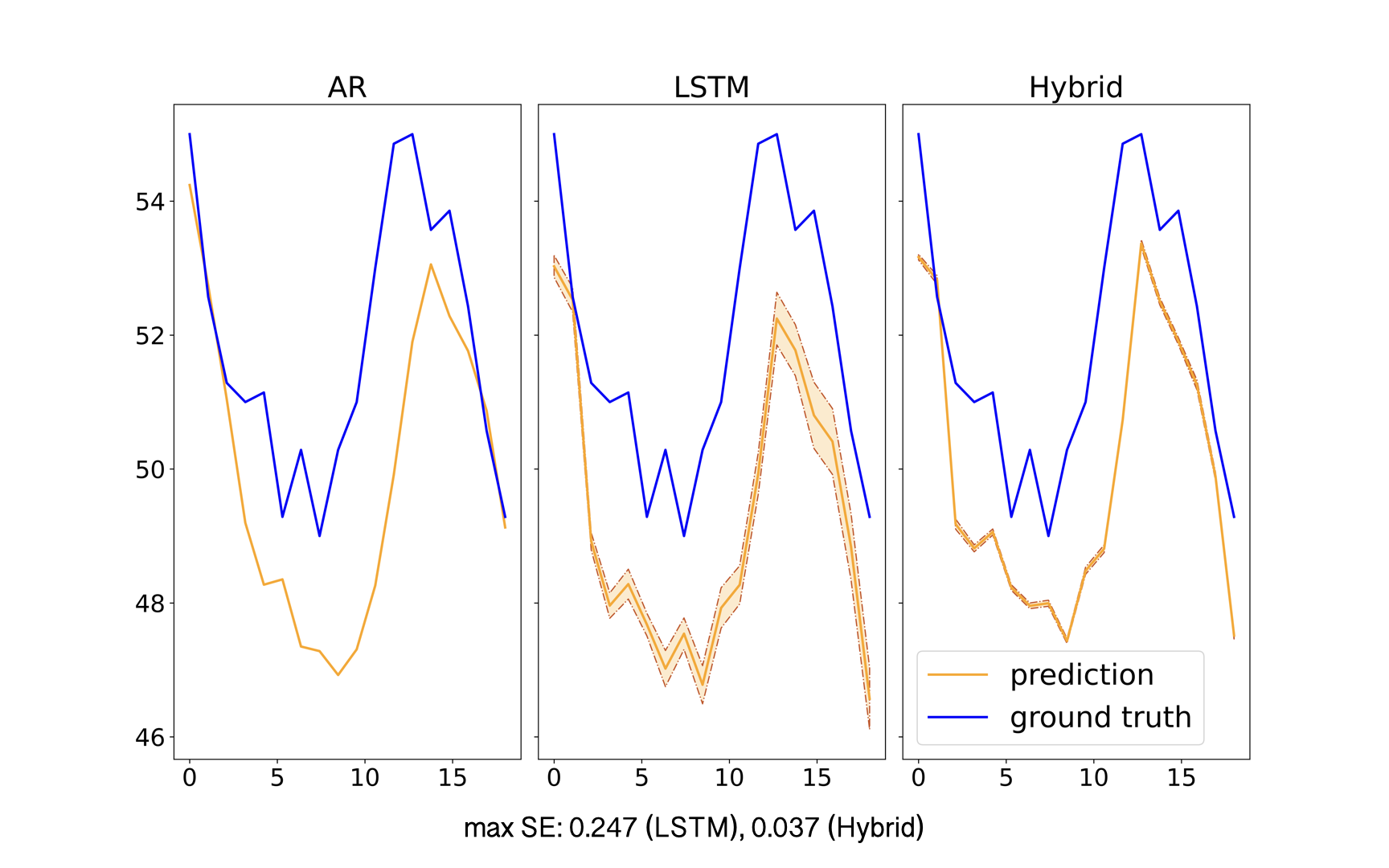}
        \subcaption{Case 6. Jagged Testing}
        \label{fig:visual_case6}
    \end{subfigure}
    \caption{The left panels show the training and testing data. The right panels show the ground truth versus forecasts of the AR, LSTM, and hybrid model, respectively. We display the average prediction (solid line) with 2 times standard error (shaded region). The standard error across 100 runs are reported for LSTM and hybrid. The hybrid model is more stable than the LSTM.} \label{fig:visualization_2}
\end{figure}

\subsection*{Interpretability}
\label{subsec:interpretability}
Interpretability of hybrid models can be defined as the ability to provide insight into the relationships they have learned, as introduced by Murdoch et al.\cite{Murdoch1900654116}. The hybrid model proposed a decomposition approach to decipher the learned model underlying the data-generating mechanism, where the estimated AR model provides the easy-to-understand linear trend. On the other hand, the LSTM is able to capture the long-term and nonlinear trend in the time series data. Our hybrid model aims to strike a balance between interpretability and accuracy, enabling us to gain insights into the underlying data while still achieving high predictive performance.

In this section, we study how AR and LSTM components contribute to the hybrid model when fitting the data. Our purpose is to seek the insights into explaining why the hybrid model enjoys the better performance in general. And more importantly, we seek to use the interpretation from the fitted hybrid model to provide practical guidance to the public health policy making process.

Note that all models are trained on the normalized data as described in section Training. Consequently all figures below report predictions on the normalized scales.

\begin{figure}[H]
\centering
\begin{minipage}{.3\linewidth} \centering \small \hspace{2mm} Los Angeles, from 2021-06-21 to 2021-09-16, $\alpha = 0.817$ \end{minipage}
\begin{minipage}{.3\linewidth} \centering \small \hspace{2mm} San Diego, from 2020-12-03 to 2021-02-28, $\alpha = 0.547$\end{minipage}
\begin{minipage}{.3\linewidth} \centering \small \hspace{2mm} Riverside, from 2021-11-18 to 2022-02-13, $\alpha = 0.174$\end{minipage}
\includegraphics[width=.32\linewidth, height=0.25\linewidth
]{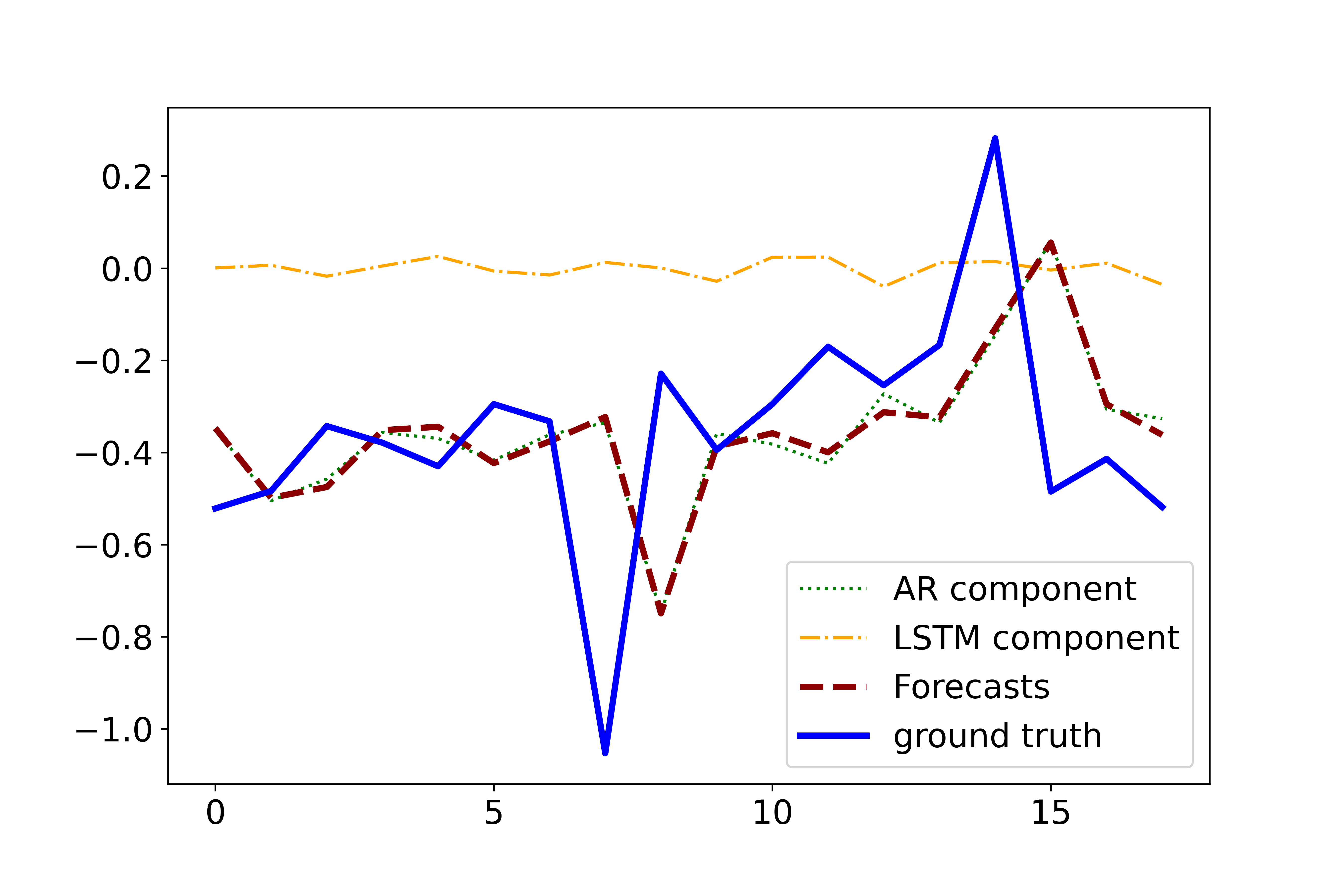}
\includegraphics[width=.32\linewidth, height=0.25\linewidth
]{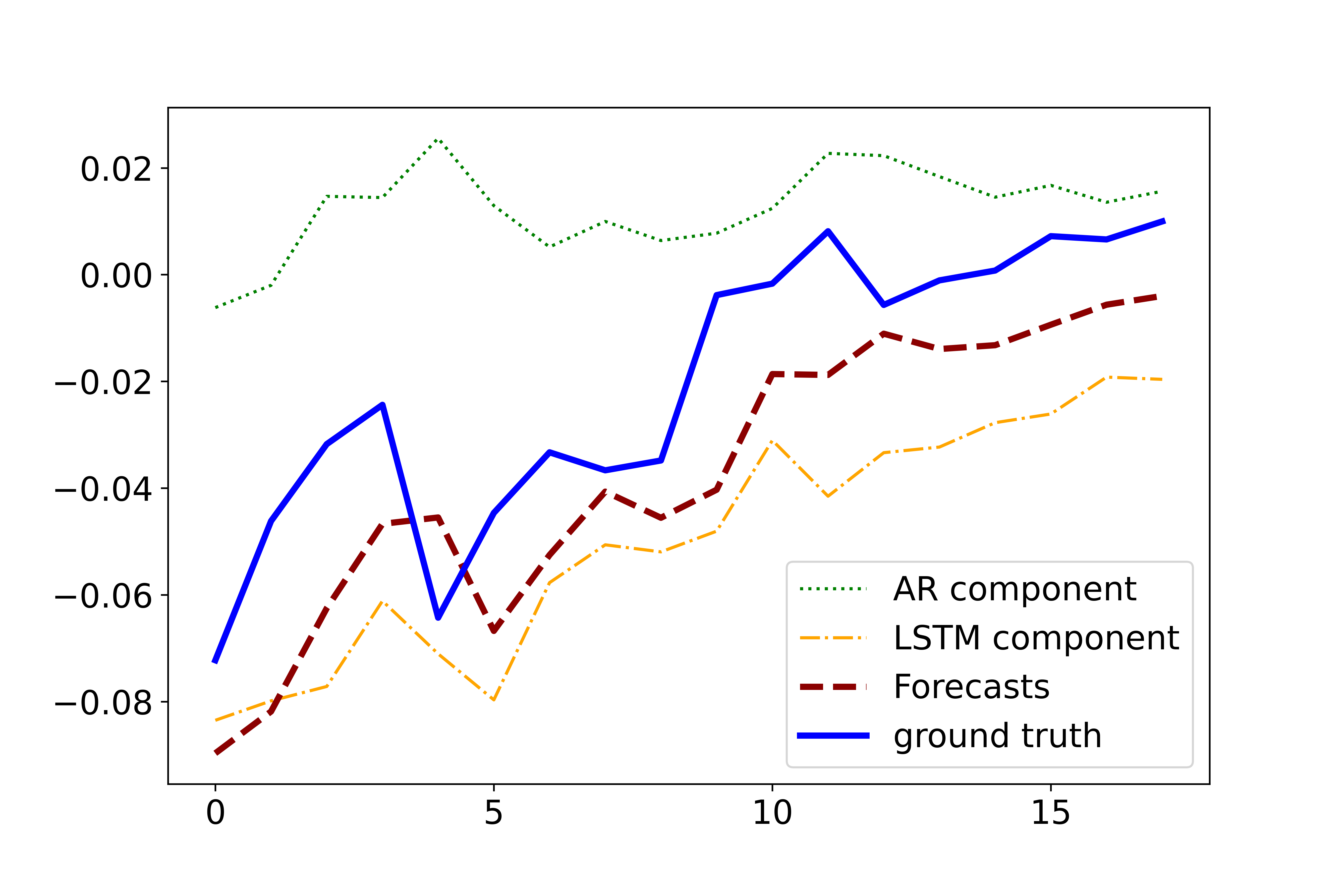}
\includegraphics[width=.32\linewidth, height=0.25\linewidth
]{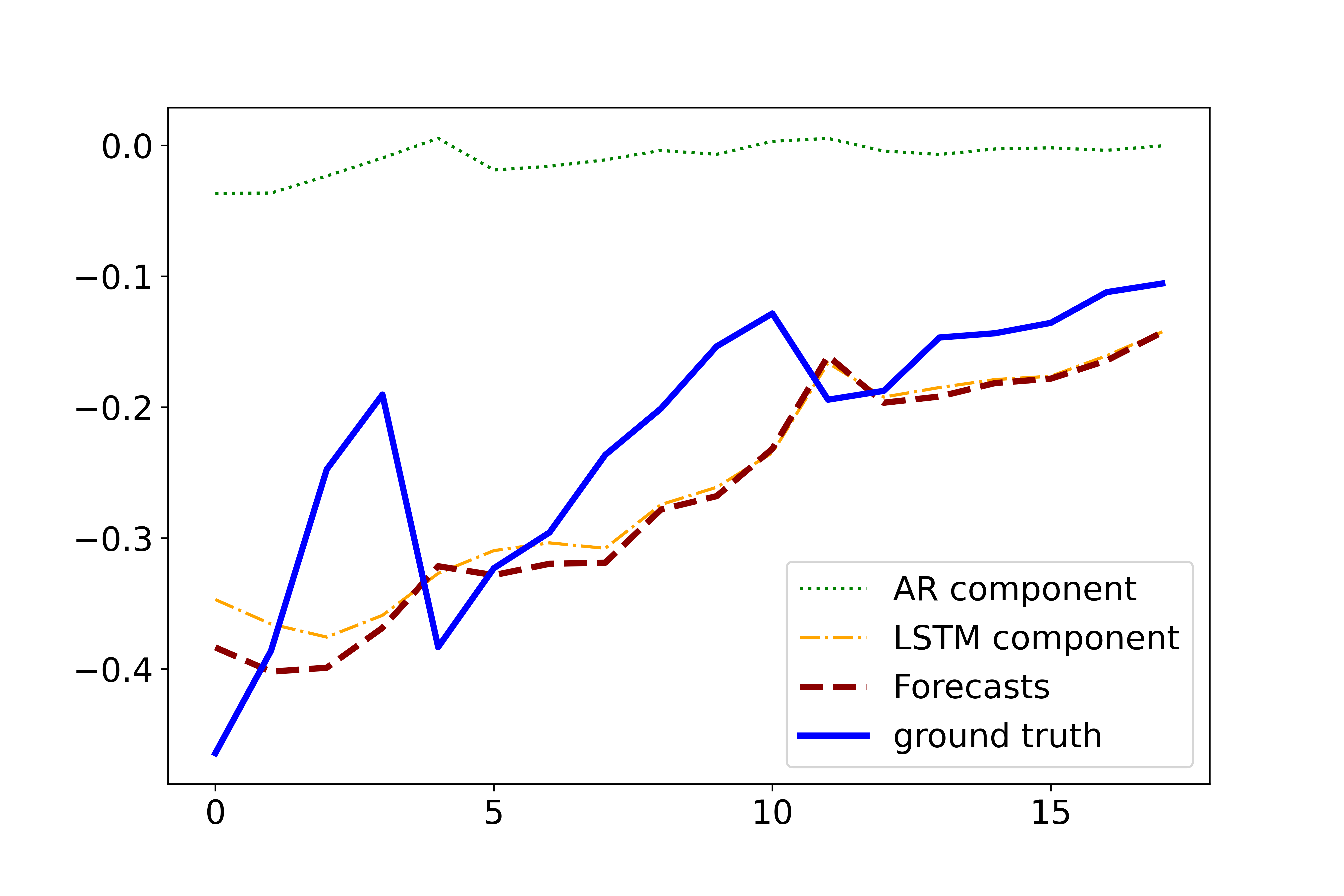}
\caption{The forecasts of a hybrid  model versus the ground truth, and the contribution of its AR and its LSTM component.}\label{fig:interpretability}
\end{figure}

In Figure~\ref{fig:interpretability}, we present three settings with different signal strength ratio (represented by the value of $\alpha$) of the AR components and LSTM components in the prediction of the hybrid model. Specifically, the larger value of $\alpha$ indicates the AR component dominates the LSTM component in prediction, and the smaller value of $\alpha$ indicates otherwise. We found that the component that has stronger signal characterizes the general trend in the data while the other helps to stabilize the variance. This observation sheds light into why the hybrid model provides better predictive performance in general than a single model.

Moreover, the fitted value of $\alpha$ provides a characterization of the intrinsic nonlinearity of the data, and consequently the difficulty of exploiting interpretation in the linear components of the fitted hybrid model. The smaller the value of $\alpha$, the higher weight the nonlinear fit using LSTM has in the final prediction. In such a setting, coefficients in the AR components should be given less weight into generating interpretation for policy making. Equivalently, for larger value of $\alpha$, it is more trustworthy to derive coefficients interpretation from the important AR part. This observation is helpful for public policy maker to distinguish among different virus transmission stages.

\begin{table}
\centering
\begin{tabular}{|l|l|l|l|l|l|l|l|l|l|}
\hline Alpha & Models & intercept & $Y_{t-1}$ & $Y_{t-2}$ & $Y_{t-3}$ & $Y_{t-4}$ & $Y_{t-5}$ & $Y_{t-6}$ & $Y_{t-7}$ \\ \hline
\multirow{2}{*}{$\alpha = 0.817$} & AR  & -0.106 & -0.124 & 0.108 & 0.063 & 0.183 & 0.001 & 0.039 & \textbf{0.654} \\ \cline{2-10} 
& hybrid & -0.008 & \textbf{0.702} & 0.142 & 0.044 & 0.210 & 0.089 & 0.114 & -0.108 \\ \hline
\multirow{2}{*}{$\alpha = 0.547$} & AR  & -0.000 & -0.019 & 0.061 & -0.077 & 0.117 & -0.041 & 0.165 & \textbf{0.553} \\ \cline{2-10}
& hybrid & 0.022 & \textbf{0.478} & -0.298 & 0.080 & -0.003 & -0.169 & 0.049 & -0.243 \\ \hline
\multirow{2}{*}{$\alpha = 0.174$} & AR  & -0.046 & 0.103 & -0.166 & 0.145 & -0.168 & 0.311 & -0.170 & \textbf{0.813} \\ \cline{2-10} 
& hybrid & 0.025 & \textbf{0.680} & 0.043 & -0.079 & -0.138 & 0.080 & -0.156 & -0.063 \\ \hline
\end{tabular}
\caption{Coefficients of AR model v.s. AR coefficients of hybrid model. When AR component dominates the prediction ($\alpha = 0.817$), MAPE for AR and hybrid are 3.088\% and 2.593\%. When the AR and the LSTM components have similar weight in prediction ($\alpha = 0.547$), MAPE for AR and hybrid are 2.523\% and 1.950\%. When LSTM component dominates the prediction ($\alpha = 0.174$), MAPE for AR and hybrid are 9.337\% and 4.665\%. Coefficients with biggest absolute values are given in bold.}
\label{table:interpret}
\end{table}
\setlength{\textfloatsep}{10pt plus 1.0pt minus 2.0pt}

Finally, we observe interesting patterns of the coefficients estimates in the AR components of the hybrid model compared with the coefficients in the pure AR model. As shown in Table~\ref{table:interpret}, across the three settings of different values of $\alpha$, the pure AR model tends to put heavier weight in coefficients of larger lags, say $Y_{t-7}$. In contrast, the AR component in the hybrid model tends to focus on capturing the short history, i.e., the coefficients associated with smaller lags (e.g., $Y_{t-1}$) tend to have larger estimates. This indicates that the short history pattern in the data could be well approximated by a simple (say, linear) model, while the longer history in the data possesses more complicated nonlinear structure that requires a LSTM component to fit.

\subsection*{Comparative study on the WHO datasets}

In this section, we compare our proposed hybrid model for COVID-19 prediction with its two component models, the ARIMA and LSTM models, as well as three other commonly used models: Support Vector Machines\cite{VapnikBF00994018} (SVM), Random Forest\cite{Breiman1999} (RF), and eXtreme Gradient Boosting\cite{Chen_2016} (XGBoost). To ensure the effectiveness of our model in different application settings, we use a country-level data for this comparative study, focusing on datasets from 7 different countries collected by the World Health Organization.

We provide a brief overview of the three additional comparing methods. Support Vector Machines (SVM) \cite{Atik5823, Muhammads42979} is a machine learning model that identifies the optimal hyperplane in a high-dimensional space that maximally separates data points into different classes. An SVM applies to both classification and regression problems. SVM is know to not perform well on noisy or unbalanced data \cite{akbani2004applying,fung2001proximal}.

Random Forest \cite{GALASSO2022111779, Abdulwahab2022, computation10060086} is an ensemble learning method that constructs a multitude of decision trees. A Random Forest is very flexible and can handle complex data types. On the other hand, the Random Forests are known for their reduced interpretability, sensitivity to noise, the need for hyperparameter tuning, and potential issues with imbalanced data. These factors may impact their performance in the context of COVID-19 predictions \cite{ANTONIADIS2021107312,ARIA2021100094,Biau2016}.

Extreme Gradient Boosting (XGBoost) \cite{Abdulwahab2022, Fange056685, LUO2021104462} has shown exceptional performance in various tasks. XGBoost is an ensemble learning method based on gradient boosting trees. It is known for its efficiency, scalability, and accuracy. However, like other tree-based ensemble methods, it can be more challenging to interpret. This may make it difficult to understand the driving factors behind predictions. In addition, XGBoost can be prone to overfitting, especially with small datasets or when the hyperparameters are not tuned properly \cite{risks7020070, li01077}.

\begin{figure}[h]
    \centering
    \includegraphics[scale=0.4]{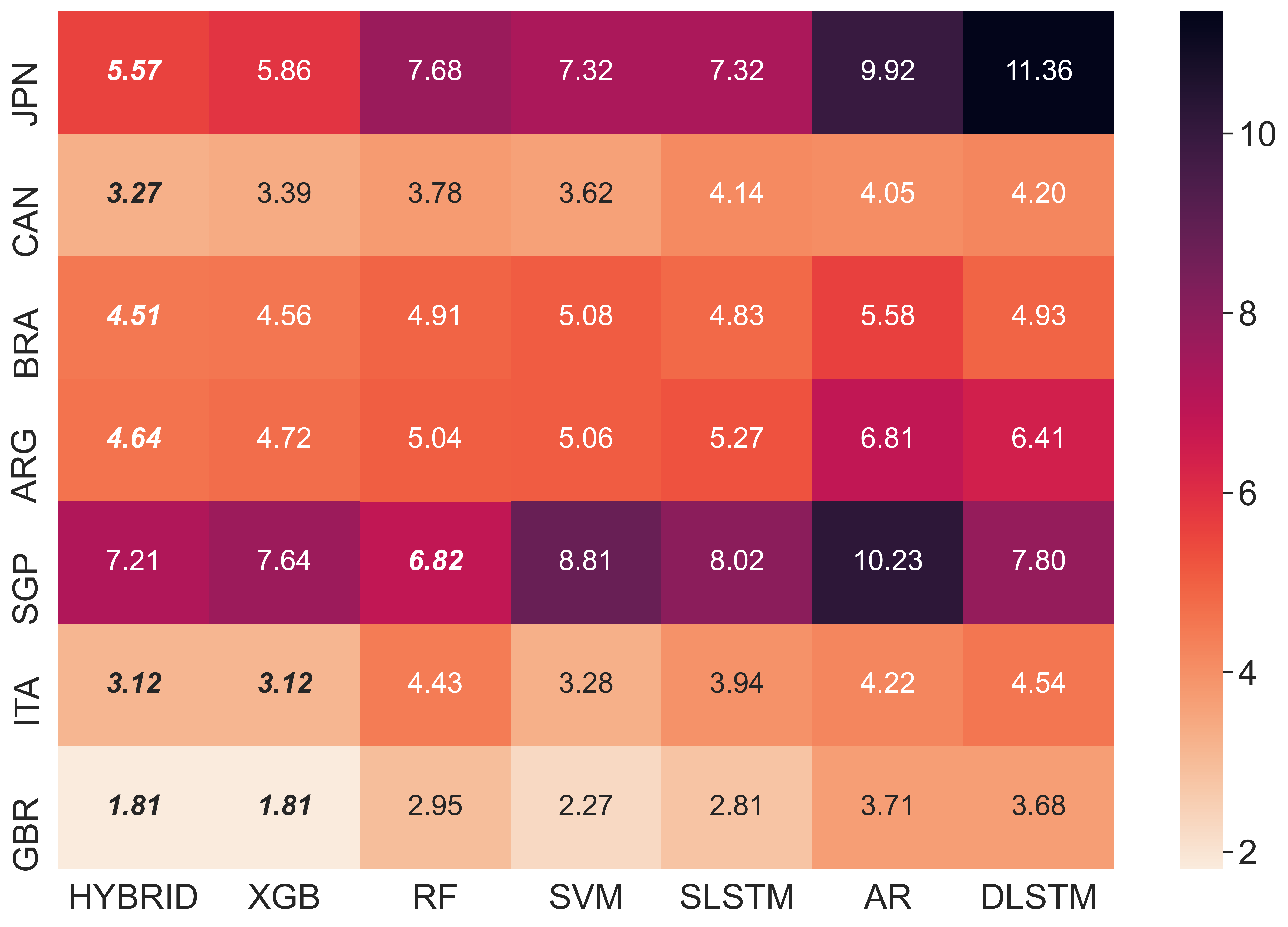}
    \caption{A heatmap exhibiting the performance, measured by MAPE in percentage, of the 7 models from this study and from previous work: AR, Single LSTM(LSTM), Double LSTM(DLSTM), hybrid, SVM, Random Forest(RF), XGBoost(XGB). The assessment has been done on data collected by World Health Organization, from 7 different countries around the world: Japan(JPN), Canada(CAN), Brazil(BRA), Argentina(ARG), Singapore(SGP), Italy(ITA), and The United Kingdom(GBR).}
    \label{fig:heatmap}
\end{figure}

We present the numerical results of the comparative study, which are visualized in Figure \ref{fig:heatmap}. The comparative study is done on data collected by the World Health Organization \cite{data_world} in Japan (JPN), Canada (CAN), Brazil (BRA), Argentina (ARG), Singapore (SGP), Italy (ITA), and the United Kingdom (GBR).

Overall, the proposed hybrid model performs better than the other models in most cases, as evidenced by its lower MAPE. This suggests that our model is effective in various situations and outperforms other commonly used models for COVID-19 prediction.

\section*{Discussion}
\label{sec:conclusion}

In this paper, we introduce a novel hybrid model that borrows strength from a highly structured Autoregressive model and a LSTM model for the task of COVID-19 cases prediction. Through intensive numerical experiments, we conclude that the hybrid model yields more desirable predictive performance than considering the AR or the LSTM counterpart alone. In principle, the hybrid model enjoy the advantages of each of its two building blocks: the expressive power of LSTM in representing nonlinear patterns in the data and the interpretability from the simple structures in AR. Consequently, the proposed hybrid model is useful in simultaneously providing accurate prediction and shedding light into understanding the transition of the virus transmission phases, and thus providing guidance to the public health policy making process.

It is also noteworthy that the predictive performance of the proposed hybrid model can be further improved by properly choosing the hyperparameters. Furthermore, while we considered LSTM as the nonlinear component in the hybrid model, it can be substituted by any other deep learning models.

\bibliography{references.bib}

\section*{Author contributions statement}

S.T. and Y.G. provided the initial idea and research plan, Y.Z. collected data, implement algorithms and performed simulations. All authors participated in the analysis and discussion of the results, and participated in the writing of the manuscript.

\section*{Funding}

Y. Zhang was partially supported by Raymond L Wilder Award  sponsored by University of California Santa Barbara and Hellman Family Faculty Fellowship. S.T. was partially supported by   Regents Junior Faculty fellowship, Faculty Early Career Acceleration grant sponsored by University of California Santa Barbara, Hellman Family Faculty Fellowship  and the NSF DMS-2111303. G.Y. was partially supported by Regents Junior Faculty fellowship, Faculty Early Career Acceleration grant sponsored by University of California Santa Barbara.

\newpage 
\section*{Supplementary material}

\paragraph{Structure of neural network.} We detail the structure of LSTM regression network and the cell gate below. 

\subsection*{California data}
\label{sec:data}

\subsubsection*{Missing data}
\noindent We study the number of daily COVID-19 cases in 8 California counties: Los Angeles, San Diego, San Francisco, Santa Barbara, Fresno, Sacramento, Ventura, and Riverside, from 2020-02-01 to 2022-09-05.\\
These data can be found on the official website of California State Government\cite{data}. 
Specifically, we collect data from the statewide COVID-19 cases deaths tests. 
Since there is only one missing value, collected in Los Angeles, we drop the observations of the same date from all counties' data. The final dataset contains consecutive time series data, collected from the 8 California counties, each with 948 observations collected in the past three years. We will use $Y_t$ to denote the number of COVID-19 cases at time $t$, where $t$ is an integer index that takes value in $\{1,2,\ldots,948\}$.\\
\subsubsection*{Statistical analysis}
\noindent Although we primarily focus on the number of daily COVID-19 cases, we have also analyzed other variables in the dataset, such as the number of reported cases, as illustrated in Supplementary figure \ref{fig:correlation.png}. By examining the correlations between these variables and the number of daily cases, we aim to provide insights for future research.

As depicted in the plot, there is a positive correlation between daily COVID-19 cases and timely data, such as daily total tests and reported cases. However, cumulative data appears to play a less significant role in predicting daily cases, with correlation values below 0.5. Among the variables that are positively related to daily cases, daily total tests have the highest correlation coefficient, even more than reported cases. On the other hand, daily deaths have the lowest positive correlation coefficient.
\begin{figure}
    \centering
    \includegraphics[width=0.7\linewidth]{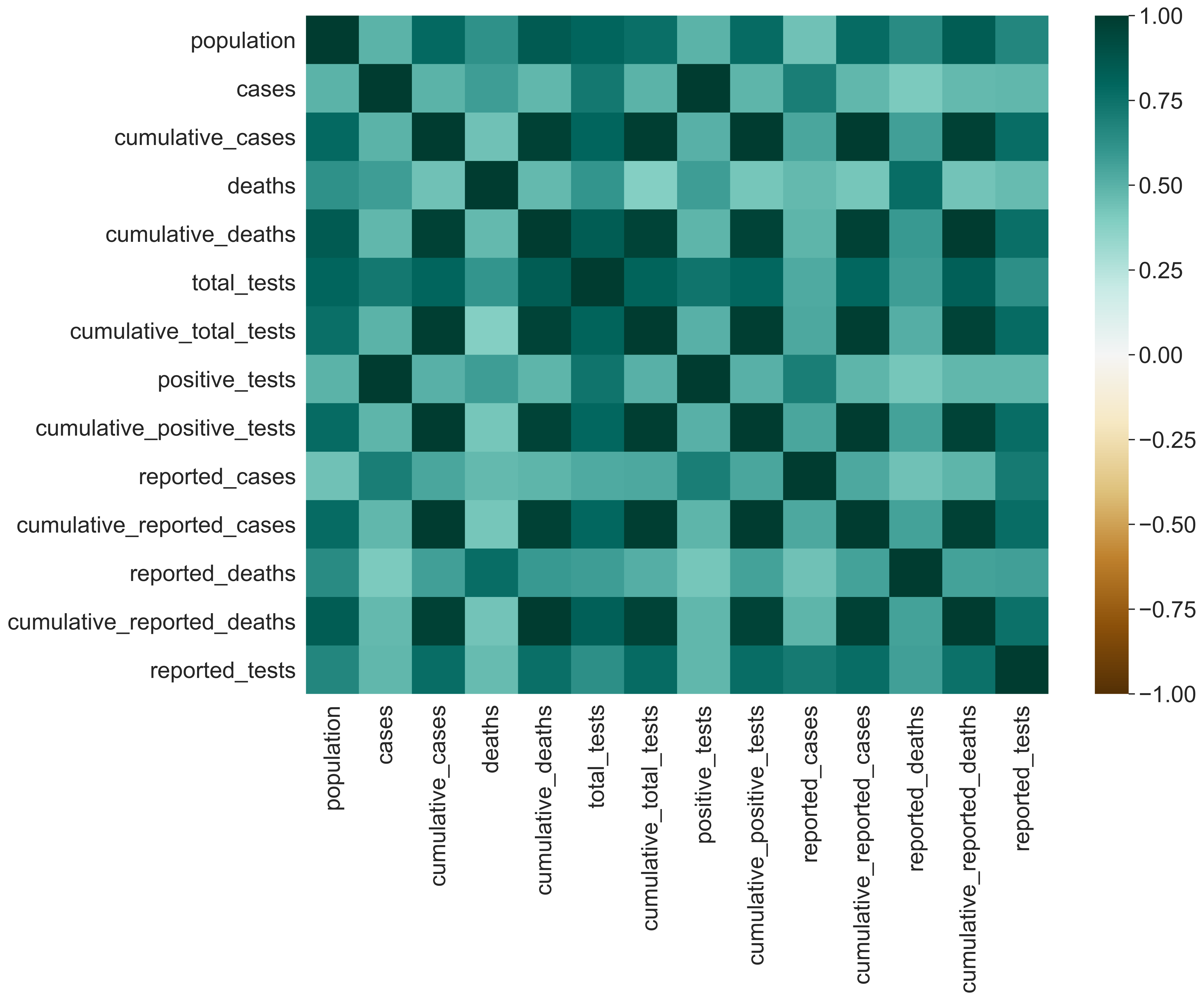}
    \renewcommand{\figurename}{Supplementary figure}
    \caption{Examining the correlation between each variable and the number of cases. The variables with darker colors are more strongly correlated with the number of cases.}
    \label{fig:correlation.png}
\end{figure}

\subsection*{Training}\label{sec:training}
In this section, we provide details on how we fit the models discussed in section Methods, and introduce the experiment setups for performance evaluation.

\subsubsection*{Data preprocessing and preparation}
 
\textbf{Smoothing}.
Recall that we denote $Y_t$ as the number of COVID-19 cases at time $t$. We adopt the standard smoothing techniques below to smooth out the irregular roughness between time steps:
$$\bar{Y}_t:=\frac{1}{\tau}\left(\sum_{i=t-\tau+1}^{t}Y_i\right),$$
where $\tau$ is the smoothing lag number.
In our study, we choose $\tau = 7$ by exploiting the day of week effect: the number of observations each date is influenced by which day of a week it is. For example, the observations of COVID-19 case detection are empirically higher on Fridays, which might be attributed to the fact that people have time to do the testing on Fridays. The day of week effect has long been observed in healthcare and the stock market. To avoid the day of week effect, we smoothed the data by using the average of observations from 6 previous dates and the current date.
We will use the smoothed data as the ground truth throughout the study.

\subsection*{Experiment setup for a single trial}

In reality, we often aim at making reliable and timely predictions based on an appropriate length of data. Furthermore, it is well-known that the different variants of COVID-19 could result in different transmission mechanisms, leading the joint distribution of confirmed cases to shift from time to time. Therefore, it is not advised to train the model using all historical data since pandemic. 
Therefore, we cut the time series into different pieces and split each piece into training set and test set. We then apply the model training and testing on each piece, which we call a trial. For each trial, we do a three-step data preprocessing:\\

\noindent\textbf{Differencing}: Statistical methods such as AR have guaranteed performance on stationary time series. However, the raw data are typically not stationary. Differencing can help us stabilize the mean of a time series by removing changes in the level of a time series, and therefore eliminating (or reducing) trend and seasonality. This can make the differenced time series stationary. Specifically, we keep differencing data until stationarity is achieved by standard tests such as Augmented Dickey Fuller Unit Root Test (ADF)\cite{ADF} or Kwiatkowski Phillips Schmidt Shin Test (KPSS)\cite{KPSS}.
We found a consistent stationarity among different trials with only one differencing operation.

\noindent\textbf{Rescale}. After differencing the data, we rescale the training data into the $[-1,1]$ with:
$$Y_{\mathrm{train, scaled}} = \frac{Y_{\mathrm{train}} - \mu_{\mathrm{train}}}{Y_{\mathrm{max, train}} - Y_{\mathrm{min, train}}}$$
where $\mu_{\mathrm{train}}$ denotes the mean of $Y_{\mathrm{train}}$, $Y_{\mathrm{max, train}}$ denotes the maximum of $Y_{\mathrm{train}}$, $Y_{\mathrm{min, train}}$ denotes the minimum of $Y_{\mathrm{train}}$, and $Y_{\mathrm{train, scaled}}$ denotes training data after scaling.  We also apply the same re-scaling map to the testing data:
$$Y_{\mathrm{test, scaled}} = \frac{Y_{\mathrm{test}} - \mu_{\mathrm{train}}}{Y_{\mathrm{max, train}} - Y_{\mathrm{min, train}}}$$
Thus the testing data does not affect the selection of our scalar. We train the model on $Y_{\mathrm{train, scaled}}$ and make predictions with $Y_{\mathrm{test, scaled}}$. Finally, we apply the inverse rescaling map to retrieve the original scale, and thus make comparison with the original data.\\

\noindent\textbf{Reshaping}. After differencing and rescaling the data, we transform the data into supervised learning form as shown below in the Supplementary table \ref{tab:supervised}. 

\begin{table}[H]
    \centering
    \begin{tabular}{rll}
    & Input & Output \\
    \hline
Row 1&$[Y_{t-n},\cdots,Y_{t-n+1}]$ & \quad $Y_{t-n+p}$ \\
Row 2&$[Y_{t-n+1},\cdots,Y_{t-n+2}]$ & \quad $Y_{t-n+p+1}$ \\
Row 3&$[Y_{t-n+2},\cdots,Y_{t-n+3}]$ & \quad $Y_{t-n+p+2}$ \\
    $\vdots$&  $\quad\quad \vdots$ & $\quad\quad \vdots$
   \end{tabular}
   \renewcommand{\tablename}{Supplementary table}
    \caption{The input-output data format. Here $p$ is the lag number. }
    \label{tab:supervised}
\end{table}

\noindent This step is also conducted on the testing data for all three models and the training data for the LSTM model and the hybrid  model.

We are now ready to conduct the experiment on a single trial. To begin with, each trial has $88$ continuous observations. We apply a first order differencing to make the trial data stationary, at the cost of losing one observation. Now we have $87$ observations. The first $62$ would be used as the training data: notice that our training size is $63$, since the $62$ differenced values are derived from $63$ observations. The remaining $25$ values will be used to make a testing data matrix.

We apply the rescaling on the training and test data. Then we reshape the testing data of length $25$ to a matrix of size $(18,8)$. For each of the $18$ rows, the first $7$ values are model inputs, which would return a single predicted value to us: an prediction for the rescaled $Y_t^{(1)}$, where the superscript 1 refers to the first order differencing. Let us denote this prediction with:
\begin{equation}
    \label{eqn:predicted_scaled}
    \widehat{Y}^{(1)}_{t, \mathrm{scaled}}
\end{equation}
We derive an prediction of the ground truth $Y_t$ from (\ref{eqn:predicted_scaled}). First, we scale (\ref{eqn:predicted_scaled}) back to the original scale by applying the inverse function of rescaling to it. Then (\ref{eqn:predicted_scaled}) becomes an prediction of  $Y_t^{(1)}$, say $\hat{Y}_t^{(1)}$. We now retrieve an estimation of the ground truth $Y_t$ with:
\begin{equation}
    \hat{Y_t} = \widehat{Y}^{(1)}_{t} + Y_{t-1}
\end{equation}
Notice that all observations before $Y_t$ are known. Since we obtain one estimation from each row, we will end up with a list of 18 estimated values. We assess the model by comparing these $18$ estimations with the $18$ corresponding ground truth values.\\

\subsection*{Choice of hyper-parameters}

In this section, we detail how we choose the hyper-parameters for each predictive model.\\

\noindent\textbf{AR}: We used lag number $7$ for the sake of interpretability, since $7$ is the number of days in a week. There could exist more scientific methods to select the lag number. For example, we may check the Bayesian information criterion (BIC). BIC is a class of information criteria to measure the goodness of fit of a statistical model. It builds on the concept of entropy and can weigh the complexity of the estimated model against the goodness of fit of this model to the data. This information helps to assess the model’s parameters and how well the model performed. \\

\noindent\textbf{LSTM}: We use batch number 1, epoch number 100, and units 1. The neural network is trained on mean square error, with optimizer adam. The fully connected layer is activated by the default linear function. Since the performance is well enough for our purpose, we do not tune the model further.\\

\noindent\textbf{Hybrid  model}: This neural network is the addition of 2 layers: an AR layer and a LSTM layer, after being weighted by a trainable coefficient between 0 and 1. We use the same set of hyperparameters as we do in the AR model and the LSTM model. As a result, we have some flexibility to tune to make the performance better.

\subsection*{Additional evaluation measures}
\subsubsection*{RMSE}
RMSE measures the difference between the forecast and the ground truth. It is a suitable metric because it removes the influence of data size and yields a result that has the same units as the input data: in other words, it is interpretable and explicit, and we can use it to compare our models to others' work, done on different datasets. RMSE is calculated by

\begin{align}
    \mathrm{RMSE} = [\sum_{i=1}^{n}(\widehat{Y}_{t} - Y_{\mathrm{true}})^2/n]^{\frac{1}{2}}
\end{align}

\subsubsection*{MAE}
MAE is another metrics commonly used to evaluate regression models. It measures the mean absolute difference between the forecasts and the ground truth values. MAE is calculated by

\begin{align}
    \mathrm{MAE} = \frac{1}{n}\sum_{i=1}^{n}|\widehat{Y}_{t} - Y_{\mathrm{true}}|
\end{align}

\noindent Just as MAPE, for both RMSE and MAE, a method with lower value is preferred.

\subsubsection*{Additional prediction results}
Beside MAPE, we evaluate and compare RMSE and MAE values of the models on the same dataset as in section Results. As shown in Supplementary table \ref{table:performance_RMSE} and \ref{table:performance_MAE}, the hybrid model has the best general performance, indicated by its lowest RMSE and MAE for each county, and it is usually the best model on the last trial. Besides, the hybrid model usually has the lowest standard error, suggesting its performance is stabler than that of its competitors.

\begin{table}
\centering
\begin{tabular}{|l|l|l|l|l|l|}
\hline
& County & AR & LSTM & LSTM (Double) & hybrid \\ \hline
\multirow{8}{*}{General performance} & Los Angeles & 265.490 (115.718) & 233.677 (91.393) & 260.866 (91.526) & \textbf{203.208} (92.471) \\ \cline{2-6} 
&San Diego & 68.809 (27.350) & 60.559 (24.826) & 67.053 (24.769) & \textbf{51.445} (20.784) \\ \cline{2-6} 
& San Francisco & 11.880 (4.441) & 10.107 (3.518) & 11.561 (3.823) & \textbf{7.943} (2.520) \\ \cline{2-6} 
& Santa Barbara & 8.926 (2.853) & 7.314 (2.257) & 8.585 (2.359) & \textbf{6.732} (2.181) \\ \cline{2-6} 
& Fresno & 18.918 (4.284) & 16.649 (4.311) & 17.964 (4.441) & \textbf{13.635} (2.988) \\ \cline{2-6} 
& Sacramento & 24.864 (9.058)  & 21.880 (8.272) & 22.844 (7.632)  & \textbf{18.183} (6.809) \\ \cline{2-6} 
& Ventura & 18.159 (5.825) & 16.219 (6.295) & 17.918 (6.137) & \textbf{13.557} (4.644) \\ \cline{2-6} 
& Riverside & 49.038 (15.131) & 41.535 (13.401) & 49.035 (14.847) & \textbf{35.246} (11.030) \\ \hline
\multirow{8}{*}{Latest performance} & Los Angeles & 93.059 & 76.594 (1.578) & 76.909 (2.473) & \textbf{74.740} (0.506) \\ \cline{2-6} 
& San Diego & 26.829 & 24.277 (0.277) & 23.987 (0.392) & \textbf{23.312} (0.088) \\ \cline{2-6} 
& San Francisco & 6.393 & 5.498 (0.057) & 5.458 (0.043) & \textbf{5.439} (0.011) \\ \cline{2-6} 
& Santa Barbara & 4.719 & 4.443 (0.031) & \textbf{4.377} (0.051) & 4.391 (0.012) \\ \cline{2-6} 
& Fresno & 8.687 & 7.765 (0.090) & 8.369 (0.374) & \textbf{7.544} (0.017) \\ \cline{2-6} 
& Sacramento & 14.179 & 11.957 (0.094) & \textbf{11.765} (0.133) & 12.269 (0.054) \\ \cline{2-6} 
& Ventura & 7.404 & 6.274 (0.038) & 6.243 (0.065) & \textbf{6.126} (0.025) \\ \cline{2-6} 
& Riverside & 23.653 & 18.311 (0.354)  & 18.228 (0.557) & \textbf{18.217} (0.210) \\ \hline
\end{tabular}
\renewcommand{\tablename}{Supplementary table}
\caption{RMSE for each model on each county. General performance is averaged on all trials. The inconsistent performances of neural networks have been compensated by the small step value, which is 7. The Latest performance is on the latest trial, from 2022-06-10 to 2022-09-05. The results for LSTM, LSTM double and hybrid are each averaged on 100 runs, to compensate the inconsistent performances of neural networks. The value in parenthesis is the standard error. AR has 0 or small standard error for the same trial. The hybrid model is usually the best in performance and has the lowest standard error.} \label{table:performance_RMSE}
\end{table}

\begin{table}
\centering
\begin{tabular}{|l|l|l|l|l|l|}
\hline
& County & AR & LSTM & LSTM (Double) & hybrid \\ \hline
\multirow{8}{*}{General performance} & Los Angeles & 208.591 (91.929) & 190.269 (76.486) & 220.444 (79.963) & \textbf{154.930} (72.476) \\ \cline{2-6} 
& San Diego & 55.252 (22.012) & 47.836 (19.699) & 55.140 (20.268) & \textbf{40.233} (16.836) \\ \cline{2-6} 
& San Francisco & 9.149 (3.235) & 8.166 (2.865) & 9.769 (3.326) & \textbf{6.074} (1.860) \\ \cline{2-6} 
& Santa Barbara & 7.069 (2.369) & 5.674 (1.806) & 7.097 (2.022) & \textbf{5.204} (1.737) \\ \cline{2-6} 
& Fresno & 15.026 (3.425) & 13.537 (3.654) & 14.733 (3.751) & \textbf{10.646} (2.342) \\ \cline{2-6} 
& Sacramento & 19.814 (7.339) & 17.362 (6.833) & 18.547 (6.431) & \textbf{13.934} (5.259) \\ \cline{2-6} 
& Ventura & 14.653 (4.692) & 12.919 (5.100) & 15.080 (5.349) & \textbf{10.804} (3.893) \\ \cline{2-6} 
& Riverside & 39.736 (12.249) & 34.055 (11.550)  & 40.854 (12.716) & \textbf{27.577} (8.649) \\ \hline
\multirow{8}{*}{Latest performance} & Los Angeles & 59.253 & 55.311 (2.053) & 52.760 (2.986) & \textbf{48.597} (0.527) \\ \cline{2-6} 
&San Diego & 16.291 & 15.600 (0.293) & 16.101 (0.769) & \textbf{14.848} (0.085) \\ \cline{2-6} 
& San Francisco & 5.119 & 4.045 (0.085) & 4.054 (0.066) & \textbf{3.918} (0.016) \\ \cline{2-6} 
& Santa Barbara & 3.346 & 2.902 (0.025) & \textbf{2.847} (0.043) & 2.897 (0.007) \\ \cline{2-6} 
& Fresno & 5.423 & 5.012 (0.130) & 5.876 (0.446) & \textbf{4.721} (0.014) \\ \cline{2-6} 
& Sacramento & 9.026  & 7.276 (0.112) & \textbf{7.118} (0.148)  & 7.371 (0.051) \\ \cline{2-6} 
& Ventura & 5.029 & 4.147 (0.040) & 4.112 (0.063) & \textbf{3.918} (0.026) \\ \cline{2-6} 
& Riverside & 18.046 & 13.123 (0.473) & 12.287 (0.647) & \textbf{11.456} (0.159) \\ \hline

\end{tabular}
\renewcommand{\tablename}{Supplementary table}
\caption{MAE for each model on each county. General performance is averaged on all trials. The inconsistent performances of neural networks have been compensated by the small step value, which is 7. The Latest performance is on the latest trial, from 2022-06-10 to 2022-09-05. The results for LSTM, LSTM double and hybrid are each averaged on 100 runs, to compensate the inconsistent performances of neural networks. The value in parenthesis is the standard error. AR has 0 or small standard error for the same trial. The hybrid model is usually the best in performance and has the lowest standard error.} \label{table:performance_MAE}
\end{table}

\end{document}